\documentclass[runningheads]{llncs}
\usepackage{graphicx}
\usepackage{tikz}
\usepackage{comment}
\usepackage{amsmath,amssymb} % define this before the line numbering.
\usepackage{color}
\usepackage{multirow}
\usepackage{multicol}
\usepackage{extarrows}
\definecolor{citecolor}{HTML}{0071bc}
\usepackage{colortbl}  % to use more color with table
\usepackage{microtype}
\usepackage[misc]{ifsym}
\usepackage[breaklinks=true,letterpaper=true,colorlinks,bookmarks=false]{hyperref}

\usepackage[accsupp]{axessibility}  % Improves PDF readability for those with disabilities.

\begin{document}
% \linenumbers
\pagestyle{headings}
\mainmatter
\def\ECCVSubNumber{2691}  % Insert your submission number here

% \title{Object Depth from Motion: \\Monocular 3D Object Detection from Videos} % Replace with your title
\title{Monocular 3D Object Detection with \\Depth from Motion \vspace{-1ex}} % Replace with your title
\begin{comment}
\titlerunning{ECCV-22 submission ID \ECCVSubNumber}
\authorrunning{ECCV-22 submission ID \ECCVSubNumber}
\author{Anonymous ECCV submission}
\institute{Paper ID \ECCVSubNumber}
\end{comment}
%******************

% \begin{comment}
\titlerunning{Monocular 3D Object Detection with Depth from Motion}
\author{Tai Wang$^{1,2}$\quad Jiangmiao Pang$^2$$^{\textrm{\Letter}}$\quad Dahua Lin$^{1,2}$}
\authorrunning{T. Wang, J. Pang, D. Lin}
\institute{$^1$The Chinese University of Hong Kong\quad $^2$Shanghai AI Laboratory\\
    \email{\{wt019,dhlin\}@ie.cuhk.edu.hk}, \email{pangjiangmiao@gmail.com}}
% \end{comment}
%******************
\maketitle

% !TEX root = ../arxiv.tex

\begin{abstract}
% % monocular 3d object detection & depth roles & previous work
% Perceiving objects in the 3D world is important for robotic systems.
% However, this has shown to be challenging for methods with only monocular input, given their ill-posed property caused by missing depth.
% Previous work mainly focuses on addressing this problem by reasoning the geometric information from a single image while ignoring the potential strong clues provided by ego-motion.
% % theoretical analysis & framework design
% In this paper, we start with a theoretical analysis for comparing the general two-view and binocular system.
% With its guidance, we propose a framework for 3D object detection from monocular videos.
% It lifts 2D features to 3D space via a depth estimation module and detects 3D objects thereon.
% The depth-from-motion system is built upon stereo estimation that relies on the construction of a geometry-aware cost volume.
% Due to its intrinsic dilemmas such as moving objects, we further compensate it with another pathway of monocular understanding.
% % pose free
% Finally, for pose-free cases in the wild, we also provide an effective formulation and a self-supervised paradigm to make the framework generalizable and extensible.
% % results
% We evaluate our framework on the KITTI benchmark and achieve new state-of-the-art with a significant improvement. Detailed quantitative and qualitative analysis further validate our theoretical discussion.

Perceiving 3D objects from monocular inputs is crucial for robotic systems, given its economy compared to multi-sensor settings.
It is notably difficult as a single image can not provide any clues for predicting absolute depth values. 
Motivated by binocular methods for 3D object detection, we take advantage of the strong geometry structure provided by camera ego-motion for accurate object depth estimation and detection.
We first make a theoretical analysis on this general two-view case and notice two challenges: 1) Cumulative errors from multiple estimations that make the direct prediction intractable; 2) Inherent dilemmas caused by static cameras and matching ambiguity.
Accordingly, we establish the stereo correspondence with a geometry-aware cost volume as the alternative for depth estimation and further compensate it with monocular understanding to address the second problem.
Our framework, named Depth from Motion (DfM), then uses the established geometry to lift 2D image features to the 3D space and detects 3D objects thereon.
We also present a pose-free DfM to make it usable when the camera pose is unavailable.
Our framework outperforms state-of-the-art methods by a large margin on the KITTI benchmark. Detailed quantitative and qualitative analyses also validate our theoretical conclusions. The code will be released at \url{https://github.com/Tai-Wang/Depth-from-Motion}.

\vspace{-1ex}
\end{abstract}
% !TEX root = ../arxiv.tex
\vspace{-1.5ex}
\section{Introduction}
\label{sec:introduction}
\vspace{-1ex}
% Mono3D but very few video Mono3D (the core challenge is depth -> video is a natural solution)
3D object detection is a fundamental task for practical applications such as autonomous driving. In the past few years, LiDAR-based~\cite{PointPillars,SECOND,VoxelNet,PointRCNN} and binocular-based~\cite{stereo1,stereo2,StereoRCNN,dsgn,liga} approaches have made great progress and achieved promising performance. In contrast, monocular methods~\cite{FCOS3D,monodle,miss3dconf,pgd} still yield unsatisfactory results as their depth estimation is naturally ill-posed. 
Although several works~\cite{monodle,miss3dconf,pgd,monoflex,monorcnn} made some attempts to tackle this problem, the current solutions still focus on digging out more geometry structures from \emph{a single image}. It is still hard for them to estimate accurate \emph{absolute} depth values.

\begin{figure}
    \centering
    \includegraphics[width=1.0\textwidth]{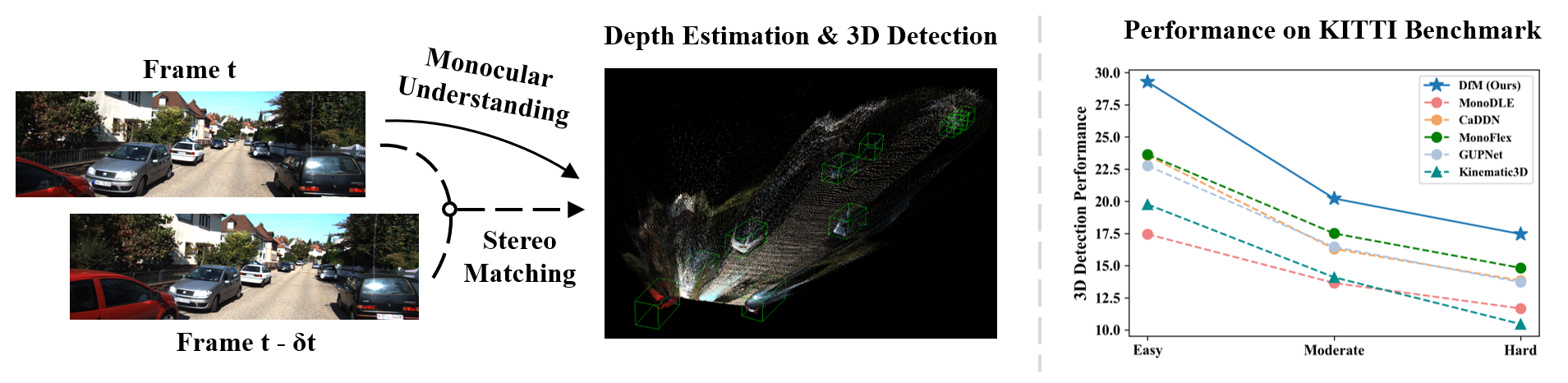}
    \vspace{-5ex}
    \caption{In this paper, we present a framework for monocular 3D detection from videos. In contrast to previous work only relying on monocular understanding from a single image, our method integrates the stereo geometric clues from temporally adjacent images. It significantly improves depth estimation accuracy, the most critical part for camera-only 3D perception, and thus enhances the 3D detection performance.}
    \vspace{-5ex}
    \label{fig:geometry}
\end{figure}
% Analyze video Mono3D (compared to 2D video, self-supervised depth, and binocular)
% This problem can be addressed if we simply extend the input from a single image to monocular videos. Monocular videos naturally provide a strong baseline prior similar to the binocular case and thus make matching-based depth estimation possible.
% The only for 3D detection~\cite{Kinematic3D} focuses on how to leverage temporal information to achieve more stable detection.
% Other methods~\cite{monodepth2,packnet} for depth estimation use the temporal adjacent image as supervision to achieve self-supervised learning without depth labels.
% Video-based 2D object detection~\cite{zhu2017vid,zhu2018vid,xiao2018vid} pays more attention to other aspects, such as occlusion and blur caused by the motion of objects.
% None of these work studies how to detect 3D objects and estimate their depth more accurately from the important ego-motion prior.
% \TOCHECK{There is some work for video depth estimation.}
This paper aims to use stereo geometry from a pair of images nearby in temporal to facilitate the object depth estimation.
The basic principle is similar to depth estimation in binocular systems. 
Two cameras in binocular systems are strictly constrained on the same plane and have a fixed distance, which is known as the system's \emph{baseline}. 
State-of-the-art stereo 3D object detection methods take this baseline as a critical clue and transform depth estimation to an easier disparity estimation problem.
Similarly, two nearby images in temporal also have stereo correspondence, but their baseline is dynamic and relies on the ego-motion of the camera. 
This idea is intuitively promising, but few previous works explored it.
The only recent work for 3D detection from monocular videos, Kinematic3D~\cite{Kinematic3D}, uses a 3D Kalman Filter and an integrated ego-motion module to build the connection between frames. It focuses on the robustness and stability of detection results but still estimates depth from a single image.
Our work, instead, is the first to study how to improve object depth estimation and 3D detection from the strong stereo geometry formed by ego-motion.

% Theoretical Analysis & Our overall framework, motivating study & technical contributions
We first conduct a theoretical analysis on this problem to better understand the geometry relationship. It reveals that direct derivation of depth in this setting involves many estimations and thus has fundamental difficulty caused by cumulative errors. The stereo estimation also has several intrinsic dilemmas, such as no baseline formed by static cameras.
% and then build a framework that detects 3D objects with depth from motion.
%Specifically, it first constructs a 3D volumetric feature from a pair of images and then performs 3D detection on its bird-eye-view transformation. % When constructing the 3D features, we use the camera pose transformation between images to apply a geometry-aware warping for calibrating the preceding features to the current frame. To guarantee the physical rationality of warping for any arbitrarily augmented input images, we technically devise a pipeline to ensure the transformation takes place in the original space, namely \emph{canonical space}.
We thus build our framework with a depth-from-motion module addressing these problems to construct 3D features and detect 3D objects thereon.
Specifically, we first involve the complex geometry relationship in a differential cost volume as the alternative for stereo estimation. To guarantee its physical rationality for any arbitrarily augmented inputs, we devise a pipeline to ensure the pose transformation takes place in the original space, namely \emph{canonical space}.
Furthermore, we compensate it with another monocular pathway and fuse them with learnable weights. 
The distribution of these learned weights well demonstrates the theoretical discussion on the intrinsic weaknesses of stereo estimation.

Considering camera poses are not always available, we also introduce a pose-free method to make the framework more flexible. 
We first decouple the ego-pose estimation as translation and rotation.
Instead of using the straightforward Euler angles, we formulate the rotation with quaternion, a more friendly representation for optimization, to avoid periodic targets.
In addition, we adopt a self-supervised loss to regularize the learning of pose to make the training get rid of pose annotations and expensive loss weights tuning.

% Experimental results
We evaluate our framework on the KITTI~\cite{KITTI} benchmark. It achieves 1st place out of monocular methods, surpassing previous methods by a large margin, 2.6\%$\sim$5.6\% and 4.2\%$\sim$7.5\% AP higher on the 3D and bird-eye-view vehicle detection benchmark respectively.
% Detailed ablation studies show the efficacy of each component and also validate our theoretical analysis.
These impressive experimental results demonstrate the potential of this stream of methods in this context, which is a more interpretable and practical perception approach like that human beings rely on.
% !TEX root = ../arxiv.tex
\section{Related Work}
\label{sec:related}
\vspace{-2ex}
\noindent\textbf{Video-Based Depth Estimation}\quad Depth estimation from monocular videos is an important problem for mobile devices and VR/AR applications. Learning-based video depth estimation methods can be divided into MVS-based (Multi-View-Stereo) methods~\cite{liu2019neural,teed2018deepv2d} and monocular-stereo hybrid methods~\cite{yoon2020novel,luo2020consistent,kopf2021robust}. The former group can not handle dynamic scenes because of the static assumption of MVS, and the latter addresses this problem by integrating a pretrained single-view depth estimator. In addition, there is another line of work~\cite{monodepth2,packnet} using videos as supervision to achieve self-supervised depth estimation. Although these works have made progress in this problem, there is still a notable gap between this field and vision-based 3D detection. Due to the disparity of scenarios and ultimate targets, previous work hardly attempts to tackle the object depth estimation problem in our context.

\noindent\textbf{Video-Based Object Detection}\quad Video-based object detection~\cite{zhu2017vid,zhu2018vid,xiao2018vid,bertasius2018vid,liu2018vid} has been studied for several years in the 2D case. These works target a better trade-off between accuracy and efficiency by aggregating features from multiple frames. Unlike the 3D case, the main problems of 2D detection from videos are the occlusion and blur of objects. The transformation between frames is generally flow-based, without considering geometric consistency in the real world. In comparison, the only previous work~\cite{Kinematic3D} for monocular 3D video object detection improves the robustness of detection results with 3D Kinematic designs. This paper is different from both. We instead focus on the specific problem in the 3D case: estimating object depth more accurately from the depth-from-motion setting and further boosting the 3D detection performance.

\noindent\textbf{Camera-Only 3D Object Detection}\quad Compared to LiDAR-based approaches \cite{PointPillars,SECOND,VoxelNet,PointRCNN,ssn,reconfig_voxels}, camera-only methods take RGB images as the only input and need to reason the depth information without accurate measurement provided by depth sensors. Among them, monocular 3D detection is more challenging than binocular because of its ill-posed property.

Earlier learning-based monocular methods~\cite{3DOP,MLFusion,Deep3DBox} used sub-networks to solve this problem. Afterward, due to the system complexity and dependence on external data and pretrained models, recent work turns to end-to-end designs~\cite{M3D-RPN,SS3D,MonoDIS,monodle,FCOS3D} like 2D detection. As several works~\cite{pgd,miss3dconf,monodle} point out the crucial rule of depth estimation in this setting, a stream of work~\cite{RTM3D,pgd,monoflex,monorcnn} attempted to address the problem with more geometric designs. Meanwhile, another line incorporates depth information to study the feature or representation transformation approaches. Pioneer work~\cite{PseudoLiDAR,OFTNet} in this line transforms the input image to 3D representations with depth estimation and performs 3D object detection thereon. Recent CaDDN~\cite{CaDDN} merges these two stages into an end-to-end framework and achieves promising results. Our work follows this high-level pipeline while focusing on improving the depth estimation from video input.

As for binocular methods, apart from the previously mentioned Pseudo-LiDAR fashion, they can be grouped into two tracks: perspective-view 2D-based~\cite{StereoRCNN,disprcnn,zoomnet,side} and bird-eye-view volume-based~\cite{dsgn,liga}. The volume-based methods are consistent with the feature transformation ideas of CaDDN. Our framework is also motivated by this stream. In contrast, we focus on studying a more difficult stereo setting: general multi-view cases formed by ego-motion.
% !TEX root = ../arxiv.tex
\begin{figure}
    \centering
    \vspace{-4ex}
    \includegraphics[width=0.95\textwidth]{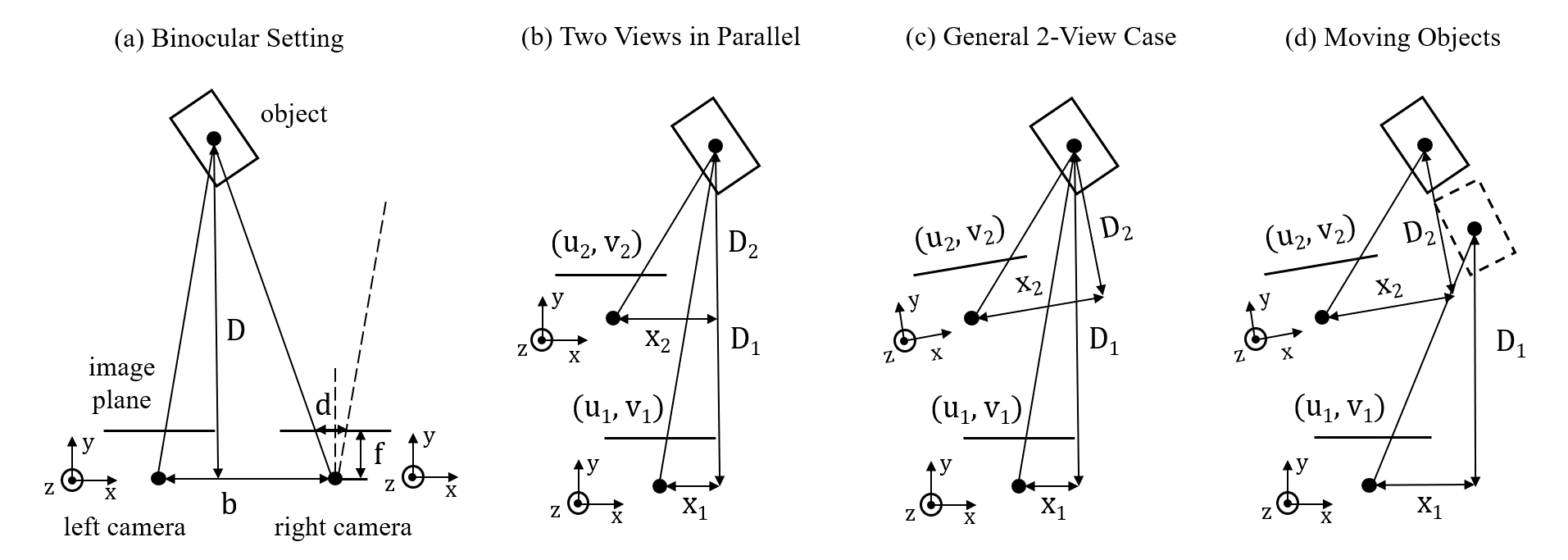}
    \vspace{-2ex}
    \caption{Multi-view geometry for object depth estimation in the (a) binocular, (b) parallel two-view, (c) general two-view system and (d) that for moving objects.}
    \vspace{-6ex}
    \label{fig:geometry}
\end{figure}
% \pjm{Please supplement figures with comprehensive captions like objects/cameras/coor. systems. Revise subfigure captions according to the captions here.}
\vspace{-2ex}
\section{Theoretical Analysis}
\vspace{-1ex}
\label{sec:theory}
In this section, we will first make a theoretical analysis for general stereo depth estimation. Among different multi-view settings, the binocular case is the simplest one and thus studied the most in the driving scenario~\cite{StereoRCNN,disprcnn,dsgn}. We start with this setting and further discuss the connection and difference when extending it to general cases. Finally, we analyze the main challenges in the depth-from-motion setting and introduce our framework design thereon.
% Finally, we take the two-parallel-view case as the example to analyze the intrinsic downsides of estimating depth from ego-motion.
\vspace{-2ex}
\subsection{Object Depth from Binocular Systems}
Binocular systems strictly constrain two cameras on the same plane.
As shown in Fig.~\ref{fig:geometry}-(a), the focal length of cameras and the distance between the pair of cameras (namely \emph{baseline} of the system) are supposed to be constant.
Following the similar triangle rule under the pinhole camera model, they follow
\vspace{-0.5ex}
\begin{equation}
    \frac{d}{f} = \frac{b}{D} \Rightarrow D = f\frac{b}{d},
    \label{eqn:simplest}
    \vspace{-0.5ex}
\end{equation}
where $d$ is the horizontal disparity on the pair of images, $f$ is the focal length of cameras, $D$ is the object depth, $b$ is the baseline.
Following Eq.~\ref{eqn:simplest}, object depth estimation can be transformed to a much easier disparity estimation problem.
%which is much easier to estimate from a pair of images.
\vspace{-2ex}
\subsection{Object Depth from General Two-View Systems}
Binocular systems rely on two-view stereo geometry to estimate object depth.
% Intuitively, two nearby images in a video also satisfy stereo correspondence.
Intuitively, two nearby images in a video also have similar stereo correspondence.
Can we use two-view geometry in this general case to predict object depth?

% We do theoretical study on this problem, and conclude that general two-view case 1) dynamic baselines changed with ego-poses and position of objects; 2) more complicated pixel-wise correspondence targets; 3) indirect relationship between depth and disparity entangled with ego rotation and object motion.

We step by step extend the geometry relationship in binocular systems to general two-view cases.
The analysis supposes the camera is in different positions at time $t_1$ and $t_2$ respectively, and we know the camera parameters at each position.
We assume all objects do not move at the beginning of this analysis and discuss the object motion at the end.

As shown in Fig.~\ref{fig:geometry}-(b), suppose the camera's movement only involves translation. We can obtain $\Delta x$ and $\Delta D$ from the transformation of camera poses.
The two-view geometry in this parallel case satisfies
\vspace{-0.5ex}
\begin{equation}
    \frac{u_1-c_u}{f} = \frac{x_1}{D_1},\quad \frac{u_2-c_u}{f} = \frac{x_2}{D_2},\quad \Delta x = x_1 - x_2,\quad \Delta D = D_1 - D_2,
    \vspace{-0.5ex}
\end{equation}
where $(u_1, v_1)$ and $(u_2, v_2)$ are a pair of corresponding points on the images, $D_1$ and $D_2$ are their depths, $x_1$ and $x_2$ are their locations in 3D space along the x-axis.
From these relationships, we can derive $D_1$:
\vspace{-0.5ex}
\begin{equation}
    D_1 = \frac{f(\Delta x-\frac{u_2-c_u}{f}\Delta D)}{u_1-u_2} \xlongequal{\Delta D=0} \frac{f\Delta x}{u_1-u_2}.
    \label{eqn:case_b}
    \vspace{-0.5ex}
\end{equation}
The geometry relationship in binocular systems is its special case when $\Delta D=0$.

As Eq.~\ref{eqn:case_b} shows, in contrast to binocular system, the "baseline" in this case is no longer fixed but dynamic that relies on camera ego-motion $\Delta x,\Delta D$ and object absolute locations $u_2$. Accordingly, object depth estimation also relies on them apart from the disparity $u_1 - u_2$.

% Thus, pixels corresponding with the same depth no longer share the same disparity between two images.

To better understand this case, we quantitatively compare it with the binocular system on KITTI as an example.
It is well-known that a suitable baseline should not be too large or small. A too-large baseline yields small shared regions of two images, while a too-small baseline results in small disparities and large estimation errors.
So we take the binocular baseline (0.54 meters on KITTI) as our example target to form with $\Delta x-\frac{u_2-c_u}{f}\Delta D$ in this case.
Because the horizontal translation $\Delta x$ is typically much smaller than 0.54 meters, we need a large translation along the depth direction ($\Delta D$) and a large horizontal distance from the 2D camera center ($u_2-c_u$) to get a baseline large enough for stereo matching. For example, to form the 0.54-meter baseline, when $\Delta D$ is 5.4 meters, $f$ is 700 pixels, then we need $u_2-c_u=70$. Accordingly, when $\Delta D$ is only 2.7 meters, then we need $u_2-c_u=140$ \footnote{For reference, the half-width of an image on KITTI is about 600 pixels.}. It means we can get more accurate estimations for objects far from central lines and may encounter problems otherwise.
%Therefore, we can observe that the basic difference of these two cases is their baselines. Because the general two-parallel-view case does not have a fixed, large enough horizontal baseline, the 3D disparity needs to be formed with two-dimension translations (horizontally and depth-wise). This observation will help us analyze the problems when estimating depth from only depth-wise translation in Sec.~\ref{sec:dfm_problems}.

% On this basis, when we also take the ego rotation and object motion into consideration, the relationship will be further entangled with them. Then we can get similar conclusions as before and more complex formulations.
On this basis, involving ego-rotation (Fig.~\ref{fig:geometry}-(c)) will introduce rotation coefficients entangled with object absolute positions to the disparity computation, and involving object motion (Fig.~\ref{fig:geometry}-(d)) will introduce \emph{relative} translation and rotation factors.
% Further involving object motion only changes the relative translation for object centers, while becoming implicit for points on the surface due to their relative rotations.
More introduction of absolute positions and motion estimation errors makes direct depth estimation more difficult. See more derivation details in the appendix.

% conclusion:
% 1. vs binocular vision (need absolute 2D position (different baselines for different pixels), baseline direction, example analysis)
% 2. hard to directly estimate the position and the disparity (and ego-motion, moving objects, and (adds-on) stereo matching problem, need mono compensation, to be discussed in the following section), essentially is still a matching problem
% 3. alternatively our modeling (need matching/correspondence relying on depth, use the modeling motivated by binocular, use the backward idea instead of forward estimate disparity? to deal with this more challenging matching problem)
% more hard matching because we have different disparities due to different 2D positions/ego-motion, instead it is the same when we have the same depth in binocular case
\subsection{Achilles Heel of Depth from Motion}\label{sec:dfm_problems}
% hard to directly estimate, more complex matching problem
% exist hard cases: no ego-motion, moving objects and stereo matching problem
% NOTE: Distinguish the concept of [depth from motion] and [general two-view system]
%       Depth from motion can also include monocular understanding in our solution
Based on the previous analysis, we can observe that direct derivation of depth in a general two-view system involves many estimations like object absolute locations and motions, thus having fundamental difficulties caused by cumulative errors.
In addition, the stereo-based solution has several cases that are intrinsically hard to handle, such as no baseline formed by static cameras and the common ambiguity problem of matching on less-textured regions.
% In addition, from the derived equation, we can anticipate several hard cases that the stereo method can not handle, such as moving objects, static cameras, and the common matching problem on less-textured regions.
% the depth-from-motion setting has challenges from two main aspects: 1) It is essentially a more complicated matching problem. Furthermore, even with decent matching solutions, we cannot estimate the depth from disparity and ego-pose with an explicit relationship. 2) From the derived equation, we can anticipate there are several hard cases that stereo methods can not handle, such as moving objects, small baselines caused by small ego-motion and the common matching problem on less-textured regions.

Therefore, motivated by binocular approaches~\cite{dsgn}, we involve the complex geometric relationship in a differential plane-sweep cost volume as the alternative to establish the stereo correspondence: Considering we can not directly estimate depth from disparity, we instead provide candidate depths for each pixel, reproject these 2.5D points to another frame and learn which one is most likely according to the pixel-wise feature similarity. Furthermore, to address the second challenge, we introduce another path for monocular understanding to compensate the stereo estimation. Next, we will elaborate on these designs with our framework in detail.

% !TEX root = ../arxiv.tex
\begin{figure}
    \centering
    \vspace{-4ex}
    \includegraphics[width=1.0\textwidth]{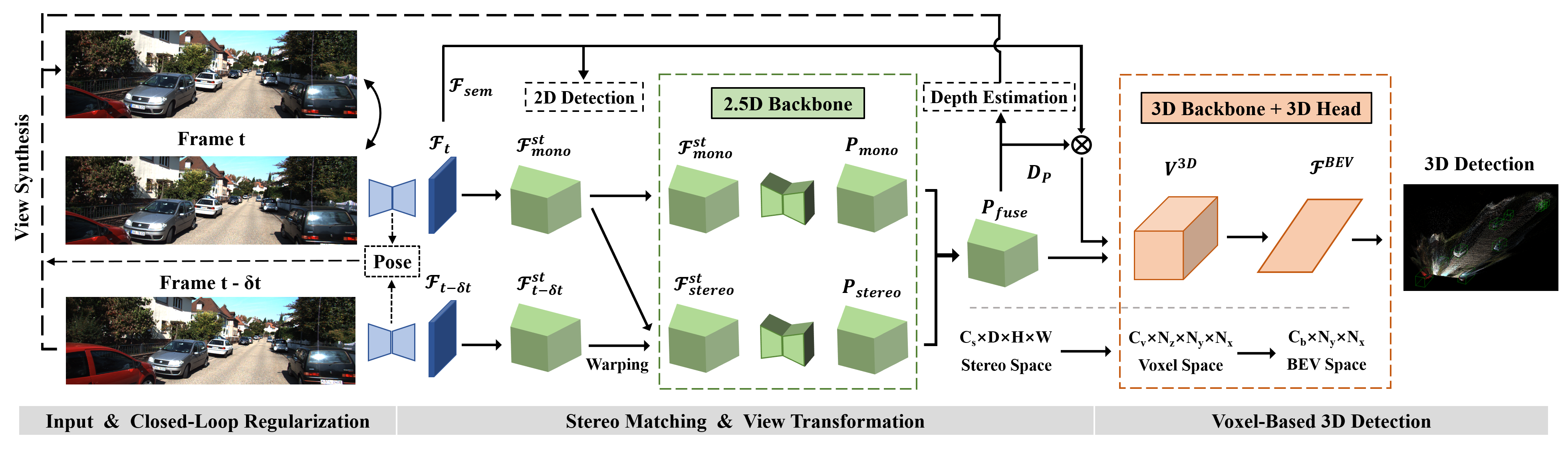}
    \vspace{-4.5ex}
    \caption{An overview of our framework.}
    \vspace{-7ex}
    \label{fig:overview}
\end{figure}
\vspace{-2.5ex}
\section{Methodology}
\label{sec:method}
\vspace{-1ex}
% framework:
% forward logic (no need for motivating study and stress on depth estimation)
% To detect 3D objects from images, depth estimation is indispensable and particularly critical to the localization accuracy. Therefore, a general pipeline for both monocular and binocular methods needs to estimate depth before performing object detection.
A general pipeline for camera-only 3D detection methods typically consists of three stages: extracting features from input images, lifting the features to 3D space, and finally detecting 3D objects thereon. We build our framework following this approach (Fig.~\ref{fig:overview}). Next, we will introduce our overall framework and present two key designs: geometry-aware cost volume construction and monocular compensation for stereo estimation. Finally, we propose a solution for pose-free cases, making the framework more flexible.
\vspace{-2.5ex}
\subsection{Framework Overview}
\vspace{-1ex}
% In this paper, we use the improved version of Deep Stereo Geometry Network (DSGN)~\misscite in LIGA-Stereo~\misscite as our baseline. It takes two images as input and finally detects 3D objects on the bird-eye-view (BEV). We replace the input binocular images with temporally consecutive frames and adapt the cost volume construction accordingly. Next, we briefly elaborate each component to make this paper self-contained. More details about network parameters can be referred to the supplementary material.
\noindent\textbf{2D Feature Extraction}\quad Motivated by binocular approaches~\cite{dsgn,liga}, given the input image-pair $(I_t, I_{t-\delta t})$, we first use a shared 2D backbone to extract their features $(\mathcal{F}_t, \mathcal{F}_{t-\delta t})$. The 2D backbone is a modified ResNet-34~\cite{ResNet} with spatial pyramid pooling (SPP)~\cite{spp} module and feature upsampling. We append a small U-Net~\cite{unet} on top to upsample the SPP feature back into full resolution. Note that here we use two different necks to generate $F_t$ as geometric feature for stereo matching and $F_{sem}$ as semantic feature following \cite{liga}. To guarantee the semantic features can get correct supervision signals, they are also used to perform the auxiliary 2D detection.

\noindent\textbf{Stereo Matching and View Transformation}\quad After getting the features of two frames, we construct the stereo cost volume $\mathcal{F}_{stereo}^{st}$ with the pose transformation between them. In addition, we lift $\mathcal{F}_t$ with pre-defined discrete depth levels to get $\mathcal{F}_{mono}^{st}$ in stereo space for subsequent monocular understanding. A dual-path 3D aggregation network filters these two volumes to predict the depth distribution volume $D_P$. $D_P(u,v,:)$ represents the depth distribution of pixel $(u,v)$ over the depth levels. The depth prediction is supervised with projected LiDAR points. Details of cost volume construction and the dual-path feature aggregation will be presented in Sec.~\ref{sec:geometry-aware-cost} and \ref{sec: mono_compensate}.
Subsequently, we lift the semantic feature $\mathcal{F}_{sem}$ with $D_P$, combine it with geometric stereo feature $P_{stereo}$ as the final stereo feature, and sample voxel features thereon. As shown in Fig.~\ref{fig:overview}, this process transforms the feature in stereo space to voxel space, which has a regular structure and is thus more convenient for us to perform object detection.

\noindent\textbf{Voxel-Based 3D Detection}\quad Next, we merge the channel dimension and height dimension to transform the 3D feature $V^{3D}$ to bird-eye-view (BEV) space, and apply a 2D hourglass network to aggregate the BEV features. Finally, a lightweight head is appended to predict 3D bounding boxes and their categories. The training loss is composed of two parts as \cite{liga}: depth regression loss and 2D/3D detection loss. See more details about the loss design in Sec.~\ref{sec:implement_details}.

%\begin{table*}
%\scriptsize
%\vspace{-3ex}
%\caption{Basic adaptation can not work directly. The key is the accuracy of depth distribution.}
%\vspace{-1ex}
%\begin{center}
%\setlength{\tabcolsep}{3mm}{
%    \begin{tabular}{c|c|c|c|c|c|c}
%    \hline
%    \multirow{2}*{Methods} & \multicolumn{3}{c|}{AP$_{3D}$ IOU$\ge 0.7$} & \multicolumn{3}{c}{AP$_{BEV}$ IOU$\ge 0.7$}\\
%    \cline{2-7}
%    ~ & Easy & Mod. & Hard & Easy & Mod. & Hard\\
%    \hline
%    Binocular Baseline & 80.62 & 61.88 & 54.92 & 90.26 & 73.63 & 66.24\\
%    w/ gt depth dist. & 85.41 & 70.07 & 62.96 & 93.82 & 82.24 & 74.89\\
%    \hline
%    DfM Baseline & 17.41 & 12.93 & 11.60 & 24.78 & 18.21 & 16.06\\
%    w/ gt depth dist. & 76.70 & 63.01 & 55.74 & 87.47 & 76.62 & 69.09\\
%    \hline
%    \end{tabular}
%
%end{center}
%label{tab: motivating_study}
%vspace{-5.0ex}
%end{table*}

%\noindent\textbf{Motivating Study}\quad When we apply this adapted baseline to this video case at the beginning, it turns out that the detection performance drops precipitously. However, if we replace the predicted depth distribution $\hat{D}_P$ with its target $D_P$, our baseline can be directly lifted to a level comparable with the binocular case. Although this assumption is a little idealistic, it still indicates that the key problem of this large gap is the accuracy of depth estimation. Therefore, we focus on improving the stereo depth estimation component next and propose two effective designs.
\begin{figure}
    \centering
    \vspace{-5.5ex}
    \includegraphics[width=1.0\textwidth]{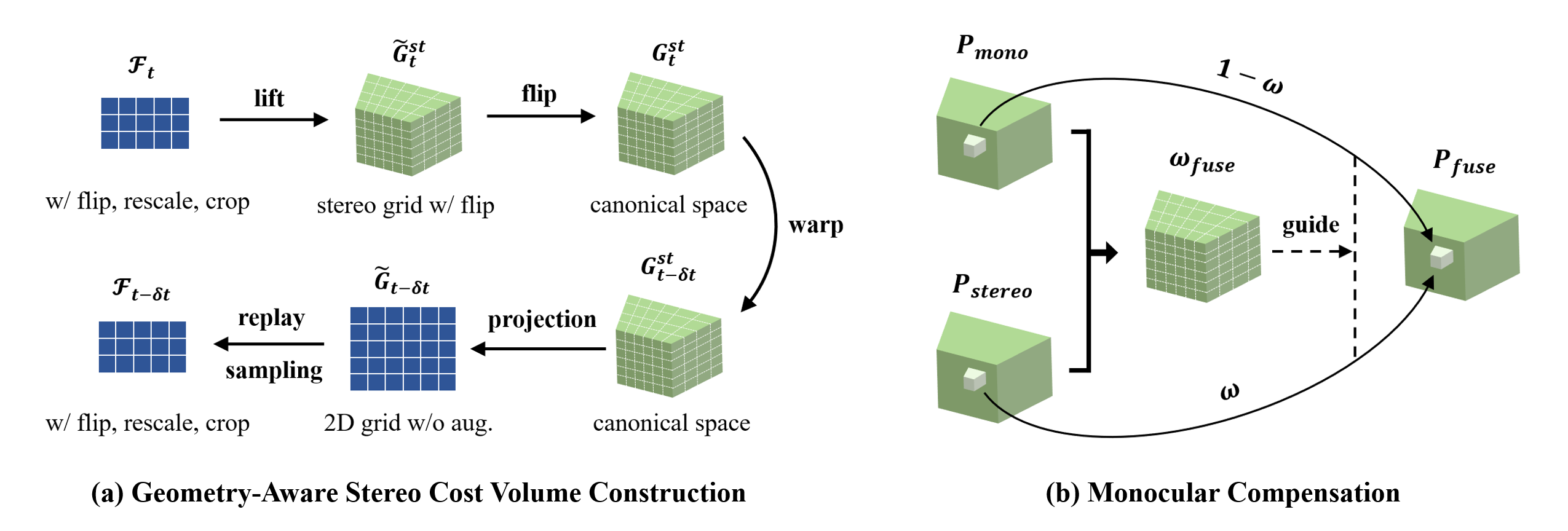}
    \vspace{-5ex}
    \caption{Key components in our depth-from-motion module.}
    \vspace{-6ex}
    \label{fig:geometry-aware-aug-fuse}
\end{figure}
\vspace{-2ex}
\subsection{Geometry-Aware Stereo Cost Volume Construction}\label{sec:geometry-aware-cost}
\vspace{-0.5ex}
% As mentioned previously, one of the key components in the framework is the construction of stereo cost volume.
The key component in the previously mentioned stereo matching is the construction of cost volume.
In contrast to the binocular case, the pose transformation between two frames is a rigid transformation composed of translation and rotation. This difference affects the method to construct cost volume and makes it hard to perform data augmentation on input images. Next, we will first formulate the procedure of volume construction and then present how we make it compatible with arbitrarily augmented input.

Formally, for each position $\mathbf{x}=(u,v,w)$ in the stereo volume, we can derive the reprojection matrix $\mathcal{W}$ to warp $\mathcal{F}_{t-\delta t}$ to the space of frame $t$ and concatenate the corresponding feature together:
\vspace{-1.0ex}
\begin{equation}
    \mathcal{F}_{stereo}^{st}(u_t,v_t,w_t) = concat\Big [\mathcal{F}_t(u_t,v_t), \mathcal{F}_{t-\delta t}(u_{t-\delta t},v_{t-\delta t}) \Big],
    \vspace{-0.5ex}
\end{equation}
\begin{equation}\vspace{-0.5ex}
    (u_{t-\delta t},v_{t-\delta t},d(w_{t-\delta t}))^T = \mathcal{W}(u_t,v_t,d(w_t))^T,\quad \mathcal{W}=KTK^{-1}.
\end{equation}
Here $(u_t,v_t,w_t)$ and $(u_{t-\delta t},v_{t-\delta t},w_{t-\delta t})$ represent the queried pixel coordinates in the stereo space of two frames. $d(w)=w\cdot \Delta d+d_{min}$ is the function to calculate the corresponding depth, where $\Delta d$ is the divided depth interval and $d_{min}$ is the minimum depth of detection range. $\mathcal{W}$ is the reprojection matrix, which is derived by multiplying intrinsic matrix $K$, ego-motion (rigid transformation) $T$ and $K^{-1}$, assuming the intrinsic matrix does not change across two frames.
Note that the matrix-multiplication-based reprojection and indexing introduce little computational overhead upon binocular methods. An empirical comparison is provided in the appendix. We find that any data augmentation, such as image rescale or flip, can affect the physical rationality of reprojection matrix $\mathcal{W}$. Constructing a \emph{geometry-aware} cost volume from augmented images here is not as trivial as in previous camera-only detection methods.

%Revisiting our stereo depth estimation module, we observe that the first main problem is that we abandon all the data augmentations in our baseline considering the physical rationality of transformation between two frames.
Therefore, we devise an approach to addressing this problem. As shown in Fig.~\ref{fig:geometry-aware-aug-fuse}-(a), we need to find the corresponding features between a pair of augmented image features $(\mathcal{F}_t, \mathcal{F}_{t-\delta t})$. Our key idea is to guarantee the warping transformation is conducted in the 3D real world, namely \emph{canonical space}. For example, if we perform flipping, rescaling, and cropping on the input two images, we first need to append pre-defined depth levels to each 2D grid coordinate of $\mathcal{F}_t$ and lift each 2.5D coordinate to 3D. During transformation, the effect of \emph{intrinsic} augmentations like rescaling and cropping should be removed through the manipulated\footnote{Rescaling and cropping correspond to the manipulation of focal length and camera centers proportionally.} intrinsic matrix $K$. Afterward, we flip the stereo grid $\tilde{G}^{st}_t$ to get $G^{st}_t$ in the canonical space. With the recovered grid, we can further perform the pose transformation to get $G^{st}_{t-\delta t}$, project it to the 2D plane and obtain several $G_{t-\delta t}$ grid maps. Finally, we replay the image augmentations and sample the corresponding features.

In this way, we can exploit any data augmentation to the input images without influencing the intrinsic rationality of ego-motion transformation. Compared to the tricky image swapping for flip augmentation in the binocular case and other alternatives, our method is also generalizable for other multi-view cases.
\vspace{-2ex}
\subsection{Monocular Compensation}\label{sec: mono_compensate}
\vspace{-1ex}
The underlying philosophies of stereo and monocular depth estimation are different: stereo estimation relies on matching while monocular estimation relies on the semantic and geometric understanding of a single image and data-driven priors. As analyzed in Sec.~\ref{sec:dfm_problems}, there are multiple cases that stereo estimation approaches can not handle. Therefore, we incorporate the monocular contextual prior to compensate stereo depth estimation.

Specifically, as shown in Fig.~\ref{fig:geometry-aware-aug-fuse}-(b), we use two 3D hourglass networks to aggregate monocular and stereo features separately. The network for monocular path shares the same architecture with the other, except for the input channel is half given the $\mathcal{F}^{st}_{mono}$ is half of $\mathcal{F}^{st}_{stereo}$. Then we have two feature volumes $P_{mono}$ and $P_{stereo}$ in the stereo space with the same shape. To aggregate these two features, we devise a simple yet effective and interpretable scheme. First, $P_{mono}$ and $P_{stereo}$ are concatenated and fed into a simple 2D convolutional network composed of 1$\times$1 kernel, and aggregated along the depth channel, \emph{e.g.}, compressed from $2D$ channels to $D$. Then the sigmoid response of this feature serves as the weight $\omega_{fuse}$ for guiding the fusion of $P_{mono}$ and $P_{stereo}$. Formally, denoting the convolutional network as $\phi$, this procedure is represented as follows:
\vspace{-1.0ex}
\begin{equation}
    \omega_{fuse}=\sigma(\phi(P_{mono},P_{stereo})),\quad P_{fuse}=\omega_{fuse}\circ P_{stereo}+(1-\omega_{fuse})\circ P_{mono}
\end{equation}
Here $\sigma$ denotes the sigmoid function, and $\circ$ refers to element-wise multiplication. The derived stereo feature $P_{fuse}$ is directly used to predict the depth distribution after a softmax and also fed into the subsequent networks for 3D detection.

This design is clean yet effective, as to be shown in the ablation studies of Sec.~\ref{sec:ablation}. Furthermore, it is interpretable both intuitively and empirically. The weight distribution of each position on the image is derived from monocular and stereo depth distributions of the same position. It is location-aware for different regions on the image, agnostic to specific reasons of inaccurate stereo estimation, and self-adaptive to different input cases. We can also validate this expected behavior by visualizing the weight $\omega_{fuse}$ and observe where stereo or monocular estimation is more reliable. See more visualization analysis in Sec.~\ref{sec:qualitative}.

\vspace{-2ex}
\subsection{Pose-Free Depth from Motion}
\vspace{-1ex}
Now we have an integrated framework for estimating depth and detecting 3D objects from consecutive-frame images. In the framework, ego-pose serves as a critical clue like the baseline in the binocular case. We essentially estimate the metric-aware depth given the metric-aware pose transformation. Although it can be easily obtained in practical applications, here we still propose a solution for the pose-free case. It is useful for mobile devices in the wild and necessary for evaluating our final models on the KITTI~\cite{KITTI} test set.

A key to learning pose is its target formulation. It is well known that any rigid pose transformation can be decomposed as translation and rotation. Both have three Degrees of Freedom (DoF). Previous work~\cite{Kinematic3D,monodepth2} typically regresses the three-dimension translation and three Euler angles. The regression of translation $\mathbf{t}$ is straightforward. For rotation estimation, instead of estimating the periodic Euler angles, we represent the rotation target with a unit quaternion $\mathbf{q}$. It is a more friendly formulation as the network output and can also avoid the potential Gimbal Lock problem.

Therefore, the output of our pose decoder network is a 7-dimension vector, including 3-dimension translation and 4-dimension unnormalized quaternion. We use the shared encoder with the backbone and add a decoder following the design of \cite{monodepth2}. The decoder consists of a bottleneck layer and three convolutional layers. Our baseline further supervises the output with L1 loss as follows:
\vspace{-1ex}
\begin{equation}
    \mathcal{L}_t=||\mathbf{t}-\hat{\mathbf{t}}||_1,\quad \mathcal{L}_r=||\mathbf{q}-\frac{\hat{\mathbf{q}}}{||\hat{\mathbf{q}}||}||_1,\quad \mathcal{L}_{pose}=\mathcal{L}_t+\lambda_r\mathcal{L}_r.
    \vspace{-0.5ex}
\end{equation}

However, this loss design has several problems:
1) we need to adjust the weight $\lambda_r$ for different cases and it is pretty expensive;
% 2) This learning is a hard regression. There is a domain gap for two 2D images to learn the 3D ego-motion;
2) There is a domain gap for two 2D images to directly regress the 3D ego-motion;
3) It is not totally pose-free as we still need pose annotations during training.
Therefore, we use a self-supervised loss~\cite{monodepth2,packnet} to replace it, considering its strength in these aspects. Specifically, the self-supervised loss is composed of an appearance matching loss $\mathcal{L}_p$ and a depth smoothness loss $\mathcal{L}_s$:
\vspace{-1ex}
\begin{equation}
    \mathcal{L}_{pose}(I_t, I_{t-\delta t})=\mathcal{L}_p(I_t, I_{t-\delta t\rightarrow t})+\lambda_s\mathcal{L}_s
    \vspace{-0.5ex}
\end{equation}
Here $I_{t-\delta t\rightarrow t}$ represents the frame $t$ synthesized with the image and depth of frame $t-\delta t$ and the predicted pose. $\mathcal{L}_p$ and $\mathcal{L}_s$ are further defined as follows:
\vspace{-1ex}
\begin{equation}
    \mathcal{L}_p(I_t, I_{t-\delta t\rightarrow t})=\frac{\alpha}{2}(1-SSIM(I_t, I_{t-\delta t\rightarrow t}))+(1-\alpha)||I_t-I_{t-\delta t\rightarrow t}||
    \vspace{-1ex}
\end{equation}
\begin{equation}
    \mathcal{L}_s(\hat{D}_t)=|\delta_x\hat{D}_t|e^{-|\delta_xI_t|}+|\delta_y\hat{D}_t|e^{-|\delta_yI_t|}
    \vspace{-0.5ex}
\end{equation}
$\mathcal{L}_p$ is formed by the Structural Similarity (SSIM)~\cite{SSIM} term and the L1 pixel-wise loss term. $\mathcal{L}_s$ is used to regularize the predicted depth map $\hat{D}_t$ on texture-less low-image gradient regions (with small $\delta_x$ and $\delta_y$). We also follow \cite{packnet} on auto-mask techniques and hyper-parameter settings ($\alpha=0.85$ and $\lambda_s=0.001$). Note that in contrast to \cite{monodepth2,packnet}, we use the LiDAR signal to supervise the learning of depth directly and only use the self-supervised loss to learn pose. In this way, because the learning of depth is supervised by absolute depth values, we can also learn a metric-aware pose even without explicit pose annotations.

% !TEX root = ../arxiv.tex
\vspace{-2ex}
\section{Experimental Setup}
\label{sec:experimental_setup}
\vspace{-1ex}
\subsection{Dataset \& Evaluation Metrics}
\vspace{-1ex}
We evaluate our method on the KITTI dataset~\cite{KITTI}, a popular benchmark for 3D object detection. It consists of 7481 frames for training and 7518 frames for testing. The training set is generally divided into 3712/3769 samples as training/validation splits. The dataset provides images, calibration information, LiDAR points and annotations for 3D detection. For the setting in this paper, apart from these information of the current frame, we also use three temporarily preceding frames. Related pose information is extracted from the raw data following Kinematic3D~\cite{Kinematic3D}. We only use images and pose information for these preceding frames and only use LiDAR as depth supervision during training.

KITTI uses Average Precision (AP) for 3D object detection evaluation. For cars, it requires a 3D bounding box overlap of more than 70\%, while for cyclist and pedestrian, it requires more than 50\%. Following \cite{MonoDIS}, we report the $AP_{40}$ results both on the validation and testing set, corresponding to the AP of 40 recall points, which is more stable and fair for comparison.

\vspace{-2ex}
\subsection{Implementation Details}\label{sec:implement_details}
\vspace{-1ex}
\noindent\textbf{Network Details}\quad We provide details about the architecture in the appendix. As for loss design, we use a focal loss for depth supervision following \cite{CaDDN} and a detection loss composed of focal loss for classification, regression L1 loss, IoU loss and direction loss for localization following ~\cite{liga}. We set the foreground and background weight of depth estimation as 5 and 1, and use $\gamma=2$ in the depth focal loss. This design can make the network more focused on the depth accuracy of foreground regions. Here we regard the regions in the annotated 2D bounding boxes as foreground.

As for the detection range, we set $[2m, 59.6m]$ for \emph{Z} (depth) axis, $[-30m, 30m]$ for \emph{X} axis and $[-1m, 3m]$ for \emph{Y} (height) axis to avoid more false positives too far away. The depth range is divided into 288 levels and the voxel size is set to $(0.2m, 0.2m, 0.2m)$. We randomly select one of three temporarily preceding images together with the current frame as training input while use the earliest one during inference if not specified in experiments.

\noindent\textbf{Training Parameters}\quad For all the experiments, except ResNet backbone pretrained on ImageNet, we trained randomly initialized networks from scratch following end-to-end manners. The network is trained using AdamW~\cite{adamw} optimizer, with $\beta_1=0.9$, $\beta_2=0.999$. We use 8 GPUs with 1 training sample on each to train the model for 60 epochs. The learning rate is set to 0.001 for the first 50 epochs and then reduced to 0.0001. The weight decay is set to 0.0001.

\noindent\textbf{Data Augmentation}\quad As presented in Sec.~\ref{sec:geometry-aware-cost}, we can apply any kind of data augmentation to input images with the canonical space as the bridge. In practice, we exploit image flip and resize augmentation in turn, and the resize range is set to $[0.95, 1.05]$. Subsequently, we fix the input image size to 320$\times$1248 by cropping the upper part which does not contain any object. Note that we only apply the corresponding augmentation in 3D space for flip, and instead manipulate the intrinsic matrix for image rescaling and cropping.

\begin{table*}
\scriptsize
\vspace{-4ex}
\caption{$AP_{40}$ results on the KITTI validation benchmark.}
\vspace{-4ex}
\begin{center}
\setlength{\tabcolsep}{2.5mm}{
    \begin{tabular}{c|c|c|c|c|c|c|c}
    \hline
     \multirow{2}*{Methods} & \multirow{2}*{Venue} & \multicolumn{3}{c|}{AP$_{3D}$ IoU$\ge 0.7$} & \multicolumn{3}{c}{AP$_{BEV}$ IoU$\ge 0.7$}\\
    \cline{3-8}
    ~ & ~ & Easy & Mod. & Hard & Easy & Mod. & Hard\\
    \hline
    MonoDIS~\cite{MonoDIS} & ICCV 2019 & 11.06 & 7.60 & 6.37 & 18.45 & 12.58 & 10.66 \\
    MonoPair~\cite{MonoPair} & CVPR 2020 & 16.28 & 12.30 & 10.42 & 24.12 & 18.17 & 15.76\\
    MoVi3D~\cite{MoVi3D} & ECCV 2020 & 14.28 & 11.13 & 9.68 & 22.36 & 17.87 & 15.73\\
    MonoDLE~\cite{monodle} & CVPR 2021 & 17.45 & 13.66 & 11.68 & 24.97 & 19.33 & 17.01\\
    PGD~\cite{pgd} & CoRL 2021 & 19.27 & 13.23 & 10.65 &  26.60	& 18.23 & 15.00\\
    CaDDN~\cite{CaDDN} & CVPR 2021 & 23.57 & 16.31 & 13.84 & - & - & -\\
    MonoFlex~\cite{monoflex} & CVPR 2021 & 23.64 & 17.51 & 14.83 & - & - & -\\
    MonoRCNN~\cite{monorcnn} & ICCV 2021 & 16.61 & 13.19 & 10.65 & 25.29 & 19.22 & 15.30\\
    GUPNet~\cite{gupnet} & ICCV 2021 & 22.76 & 16.46 & 13.72 & 31.07 & 22.94 & 19.75\\
    DFR-Net~\cite{DFR-Net} & ICCV 2021 & 19.55 & 14.79 & 11.04 & 26.60 & 19.80 & 15.34\\
    \hline
    Kinematic3D~\cite{Kinematic3D} & ECCV 2020 & 19.76 & 14.10 & 10.47 & 27.83 & 19.72 & 15.10\\
    \hline
    DfM w/o pose & ECCV 2022 & 26.65 & 18.49 & 15.94 & 34.97 & 25.00 & 22.00\\
    DfM w/ pose & ECCV 2022 & \textbf{29.27} & \textbf{20.22} & \textbf{17.46} & \textbf{38.60} & \textbf{27.13} & \textbf{24.05}\\
    \hline
    \end{tabular}
}
\end{center}
\label{tab: kitti_val}
\end{table*}
\vspace{-7.0ex}

\vspace{-2ex}
\section{Results}
\label{sec:results}
\vspace{-1ex}
In this section, we first analyze our main quantitative results, with a comparison with other methods on the KITTI benchmark. Then we show the visualization of aggregation weights mentioned in Sec.~\ref{sec: mono_compensate} and discuss the reliability of monocular or stereo estimation in different cases.
Finally we make detailed ablation studies for each important component in our framework to reveal their efficacy.
\vspace{-2ex}
\subsection{Quantitative Analysis}
\vspace{-1ex}
% Compared to monocular (BEV \& 3D)
\noindent\textbf{Main Results}\quad First, we compare our framework with other state-of-the-art methods on the KITTI validation benchmark (Tab.~\ref{tab: kitti_val}), considering the ego-pose information is not available on the test set. We observe a significant improvement in both 3D detection and bird-eye-view (BEV) performance, 2.6\%$\sim$5.6\% and 4.2\%$\sim$7.5\% higher than the previous best for all the difficulty levels respectively. We conjecture that the better improvement on BEV performance is caused by our paradigm of voxel-based 3D detector: it finally detects 3D objects from the bird-eye-view following \cite{SECOND,PointPillars}. In addition, even without ego-pose information, our framework still outperforms others by a notable margin. This further shows the benefits brought by temporal information and stereo estimation. Please refer to the appendix for its performance on the test set and other categories.

\noindent\textbf{Comparison with Video-Based Methods}\quad Compared to the only previous methods using video information, Kinematic3D~\cite{Kinematic3D}, our method also shows significant superiority. The reason is that Kinematic3D focuses more on the stability of detection and forecasting while our method pays more attention to depth estimation. Considering that the evaluation metric on KITTI requires particularly accurate localization for detected objects, our method naturally shows better performance on the benchmark. Note that our method is also compatible with some methods proposed in Kinematic3D. They can further improve the detection stability and efficiency of our framework and provide a natural integration with the downstream tasks such as tracking, prediction and planning.

% Compared to binocular
\noindent\textbf{Comparison with Binocular Methods}\quad Although our approach has achieved promising progress over previous monocular methods, we still observe a large gap between ours and binocular state of the art (64.7\% AP for moderate). It is partly due to intrinsic weaknesses of the depth-from-motion setting. Nevertheless, we can expect a large space for improvement as the advancement of binocular methods, from RT3DStereo~\cite{RT3DStereo} (23.3\% AP) to LIGA-Stereo~\cite{liga} (64.7\% AP).

\vspace{-4ex}
\begin{figure}
    \centering
    \includegraphics[width=1.0\textwidth]{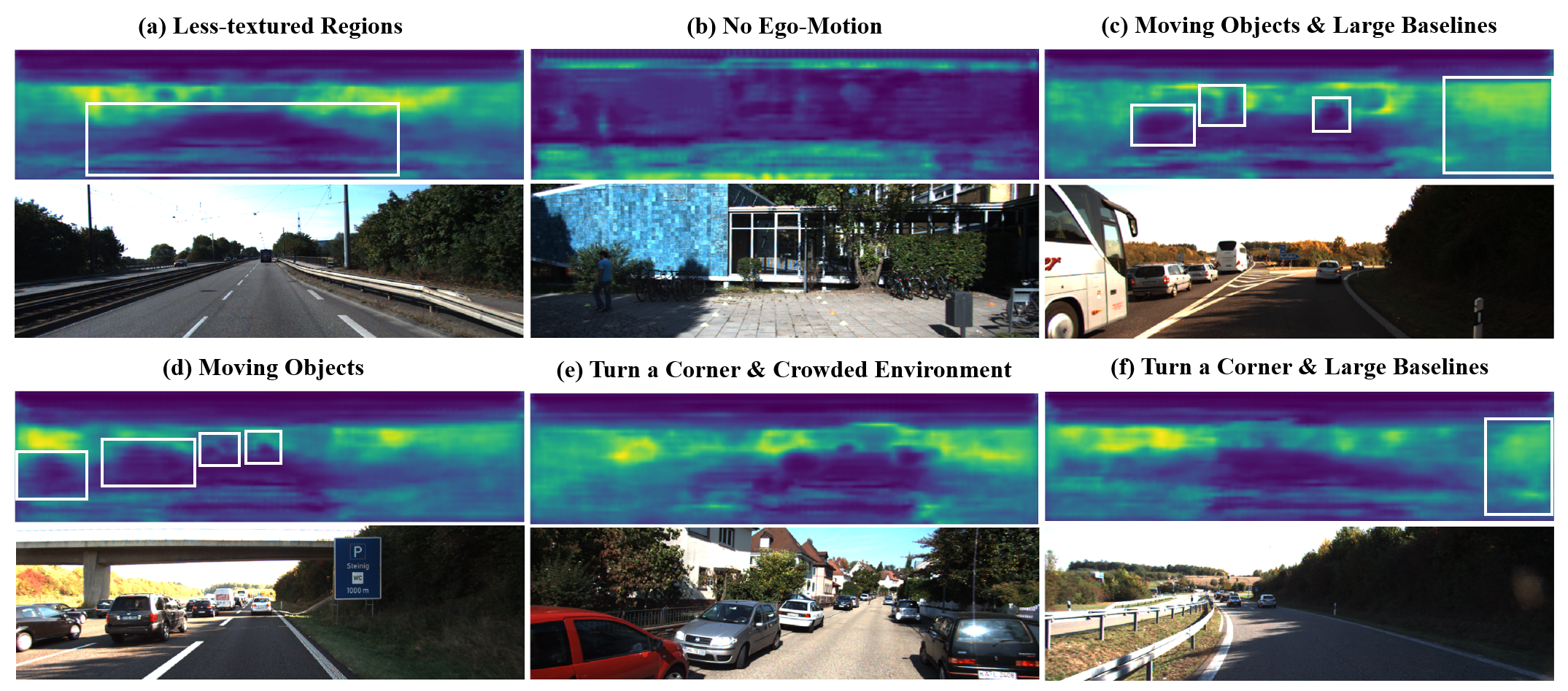}
    \vspace{-4ex}
    \caption{Qualitative Analysis of aggregation weights in different cases.}
    \vspace{-6ex}
    \label{fig:qualitative}
\end{figure}
\vspace{-2ex}
\subsection{Qualitative Analysis}\label{sec:qualitative}
\vspace{-1ex}
For qualitative analysis, we show the visualization of aggregation weights (summed along the depth axis) in Sec.~\ref{sec: mono_compensate} with some representative cases (Fig.~\ref{fig:qualitative}). For each sample plotted in the figure, we visualize the weight ranging from 0 to 1 above each image. Larger weights are marked with lighter regions in the weight maps, which indicates that the depth estimation relies more on stereo matching.

Next, we will discuss the inherent problems of stereo methods in the depth-from-motion setting analyzed in Sec.~\ref{sec:dfm_problems}. In a general case, (a) shows that the estimation relies more on monocular priors for less textured regions such as the road. (b) shows a case that stereo matching will break down: no baseline is formed by static cameras. (c) and (d) show that stereo methods can not handle moving objects with the current pure design. In addition, on the right side of image (c), when the richness of texture seems similar, the regions far away from camera centers can form larger baselines. They can thus get more accurate estimations from stereo matching. A similar phenomenon can be seen in sample (f). Finally, even the driving car is turning a corner, all of our analysis is still valid because the rotation in the ego-motion can not be quite large in a short period. This weight is also learned adaptively for the crowded environment. These prove the interpretability of our method and the necessity of monocular compensation. It also points out possible directions for improving this group of the method, such as handling moving objects with customized designs in the stereo estimation.

\begin{table}
\scriptsize
\vspace{-4ex}
\caption{Ablation studies for geometry-aware stereo cost volume construction.}
\vspace{-5ex}
\begin{center}
\setlength{\tabcolsep}{3mm}{
    \begin{tabular}{c|c|c|c|c|c|c}
    \hline
     \multirow{2}*{Methods} & \multicolumn{3}{c|}{AP$_{3D}$ IoU$\ge 0.7$} & \multicolumn{3}{c}{AP$_{BEV}$ IoU$\ge 0.7$}\\
    \cline{2-7}
    ~ & Easy & Mod. & Hard & Easy & Mod. & Hard\\
    \hline
    Baseline & 17.41 & 12.93 & 11.60 & 24.78 & 18.21 & 16.06\\
    +Flip aug. & 19.13 & 13.92 & 12.62 & 26.89 & 19.48 & 17.52\\
    +Rescale aug. & 21.47 & 15.32 & 13.83 & 29.22 & 21.22 & 19.51\\
    \hline
    \end{tabular}
}
\end{center}
\label{tab: ablation_geometry_aware}
\end{table}
\vspace{-10.0ex}
\begin{table}
\scriptsize
\caption{Detection performance of different depth estimation approaches.}
\vspace{-5ex}
\begin{center}
\setlength{\tabcolsep}{3mm}{
    \begin{tabular}{c|c|c|c|c|c|c}
    \hline
     \multirow{2}*{Methods} & \multicolumn{3}{c|}{AP$_{3D}$ IoU$\ge 0.7$} & \multicolumn{3}{c}{AP$_{BEV}$ IoU$\ge 0.7$}\\
    \cline{2-7}
    ~ & Easy & Mod. & Hard & Easy & Mod. & Hard\\
    \hline
    Mono Only & 20.06 & 15.30 & 14.05 & 27.84 & 21.78 & 19.96 \\
    Stereo Only & 21.47 & 15.32 & 13.83 & 29.22 & 21.22 & 19.51\\
    Mono+Stereo & 26.61 & 18.82 & 16.47 & 36.16 & 26.09 & 23.17 \\
    \hline
    \end{tabular}
}
\end{center}
\label{tab: det_mono_stereo}
\vspace{-10.0ex}
\end{table}

\vspace{-4ex}
\begin{figure}
    \centering
    \includegraphics[width=1.0\textwidth]{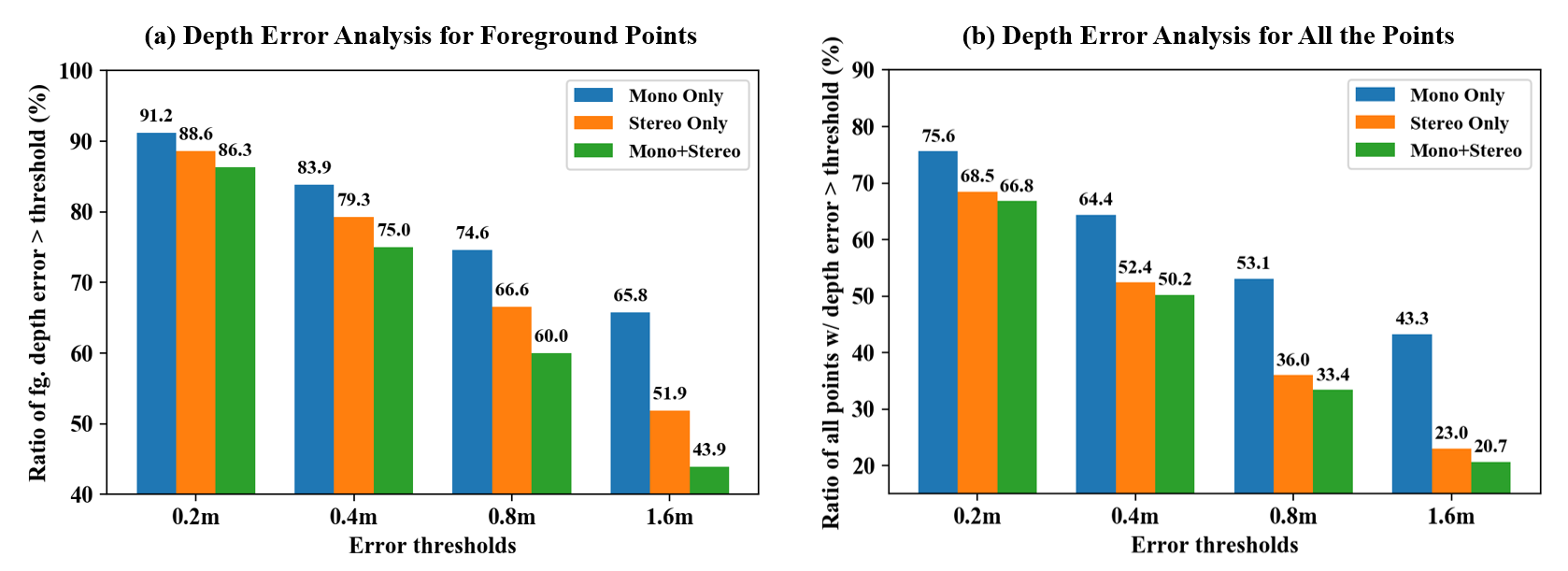}
    \vspace{-5ex}
    \caption{We make error analysis for the depth predictions of foreground region and the entire scene by different methods, respectively, by comparing the percentage of points with depth errors greater than thresholds: 0.2m, 0.4m, 0.8m, 1.6m. The error medians of monocular/stereo/hybrid methods on the foreground region/the entire scene are 5.86/3.33/2.60m and 1.15/0.58/0.48m.}
    \vspace{-6ex}
    \label{fig:depth_error}
\end{figure}

% \begin{table*}
% \scriptsize
% \vspace{-3.0ex}
% \caption{Depth estimation errors of different depth estimation approaches. Err. Med. denotes the average median of depth errors and other metrics evaluate the ratio of points with errors larger than a specific threshold. Foreground (Fg in the table) metrics are evaluated by averaging object-level results. Objects with less than 5 ground-truth LiDAR points are ignored.}
% \vspace{-1ex}
% \begin{center}
% \setlength{\tabcolsep}{2mm}{
%     \begin{tabular}{c|c|c|c|c|c}
%     \hline
%      \multirow{2}*{Methods} & \multicolumn{1}{c|}{Err. Med.$\downarrow$} & \multicolumn{1}{c|}{$>$0.2m$\downarrow$} & \multicolumn{1}{c|}{$>$0.4m$\downarrow$} & \multicolumn{1}{c|}{$>$0.8m$\downarrow$} & \multicolumn{1}{c}{$>$1.6m$\downarrow$}\\
%     \cline{2-6}
%     ~ & Fg/All (m) & Fg/All (\%) & Fg/All (\%) & Fg/All (\%) & Fg/All (\%)\\
%     \hline
%     Mono Only & 5.86/1.15 & 91.2/75.6 & 83.9/64.4 & 74.6/53.1 & 65.8/43.3 \\
%     Stereo Only & 3.33/0.58 & 88.6/68.5 & 79.3/52.4 & 66.6/36.0 & 51.9/23.0 \\
%     Mono+Stereo & 2.60/0.48 & 86.3/66.8 & 75.0/50.2 & 60.0/33.4 & 43.9/20.7 \\
%     \hline
%     \end{tabular}
% }
% \end{center}
% \label{tab: depth_mono_stereo}
% \end{table*}
\vspace{-2ex}
\subsection{Ablation Studies}\label{sec:ablation}
\vspace{-1ex}
\noindent\textbf{Geometry-Aware Stereo Cost Volume}\quad First, we show the benefits of geometry-aware stereo cost volume construction in Tab.~\ref{tab: ablation_geometry_aware}. Both flip and rescale augmentation can remarkably enhance the detector. We suspect that making the cost volume more compatible with various augmented inputs can improve the generalization ability of models for different scenes and camera intrinsic settings.

\noindent\textbf{Monocular Compensation}\quad We compare different approaches for depth estimation in Tab.~\ref{tab: det_mono_stereo} and Fig.~\ref{fig:depth_error}. We turn off one of two branches in Sec.~\ref{sec: mono_compensate} by setting the corresponding weight to zero during training and compare their detection (Tab.~\ref{tab: det_mono_stereo}) and depth estimation accuracy (Fig.~\ref{fig:depth_error}). We can see that with only monocular context, models still achieve a decent detection performance while failing on depth estimation of the entire scene. Stereo matching performs better on both aspects, especially the latter. Because these modules compensate each other fundamentally, our aggregation design brings an impressive gain thereon.

\begin{table}
\scriptsize
\vspace{-4.5ex}
\caption{Ablation studies of using different preceding frames during inference.}
\vspace{-5ex}
\begin{center}
\setlength{\tabcolsep}{4mm}{
    \begin{tabular}{c|c|c|c|c|c|c}
    \hline
     \multirow{2}*{Methods} & \multicolumn{3}{c|}{AP$_{3D}$ IoU$\ge 0.7$} & \multicolumn{3}{c}{AP$_{BEV}$ IoU$\ge 0.7$}\\
    \cline{2-7}
    ~ & Easy & Mod. & Hard & Easy & Mod. & Hard\\
    \hline
    Prev-1st & 24.09 & 17.27 & 15.03 & 35.50 & 25.24 & 22.82 \\
    Prev-2nd & 24.92 & 17.62 & 15.68 & 35.89 & 25.39 & 22.99 \\
    Prev-3rd & 25.19 & 17.96 & 15.92 & 36.16 & 25.88 & 23.03 \\
    \hline
    \end{tabular}
}
\end{center}
\vspace{-10.5ex}
\label{tab: ablation_preceding}
\end{table}

\begin{table*}
\scriptsize
\vspace{-4ex}
\caption{Ablation studies of different pose-free designs.}
\vspace{-5ex}
\begin{center}
\setlength{\tabcolsep}{3mm}{
    \begin{tabular}{c|c|c|c|c|c|c}
    \hline
     \multirow{2}*{Methods} & \multicolumn{3}{c|}{AP$_{3D}$ IoU$\ge 0.7$} & \multicolumn{3}{c}{AP$_{BEV}$ IoU$\ge 0.7$}\\
    \cline{2-7}
    ~ & Easy & Mod. & Hard & Easy & Mod. & Hard\\
    \hline
    Euler for rotation & 20.16 & 15.03 & 13.01 & 28.96 & 21.21 & 19.08 \\
    + use quaternion & 23.88 & 16.93 & 14.47 & 33.23 & 23.75 & 20.72 \\
    + use reproj. supervision & 26.65 & 18.49 & 15.94 & 34.97 & 25.00 & 22.00 \\
    \hline
    \end{tabular}
}
\end{center}
\label{tab: ablation_preceding}
\vspace{-7.5ex}
\end{table*}

\noindent\textbf{Different Preceding Frames}\quad As analyzed in Sec.~\ref{sec:dfm_problems}, the distance of ego-vehicle in two frames can affect the baseline in this depth-from-motion setting and thus affect the accuracy of stereo matching. To compare the effect of using different frames, we train the model with a randomly selected previous frame for each sample and test it with a fixed one. Note that when the sample does not have the corresponding preceding frame, for instance, the third preceding one, we will use the earliest one that it has. As Tab.~\ref{tab: ablation_preceding} shows, using the third preceding frame performs better than others up to about 1\% mAP, which validates our analysis. This study has additional space for exploration: If given more previous frames, which one would be the best choice? If we involve multiple frames into stereo matching and depth estimation, what is a better frame selection design?

\noindent\textbf{Pose-Free Designs}\quad Finally, we study the specific designs for pose-free depth from motion. Our baseline uses the Euler angle as the rotation representation as \cite{Kinematic3D} and directly regresses the translation and rotation with the pose supervision. We further try the quaternion representation and reprojected photometric loss as the supervision, and both show superiority than before. More importantly, we can avoid the pose annotation completely with the self-supervised paradigm, which is especially important for the practice in the real world.

% !TEX root = ../arxiv.tex
\vspace{-2ex}
\section{Conclusion}
\label{sec:conclusion}
\vspace{-1ex}
In this paper, we propose a framework for monocular 3D detection from videos.
% It targets improving the depth estimation with ego-motion priors and further achieves more accurate 3D localization.
% Built on a binocular baseline, we make two critical modifications on cost volume construction and monocular-compensated stereo estimation tailored to this setting.
It lifts 2D image features to 3D space via an effective depth estimation module and detects 3D objects on top. The depth-from-motion system leverages an important ego-motion clue to estimate depth from stereo matching, which is further compensated with monocular understanding for addressing several intrinsic dilemmas.
To make this framework more flexible, we further extend it to pose-free case with an effective rotation formulation and a self-supervised paradigm.
% The proposed pose-free solution further makes the framework integrated and extensible.
Experimental results show the efficacy of our method and validate our theoretical discussion.
In the future, we will optimize our framework in terms its simplicity and generalization ability. How to address the stereo estimation of moving objects is also an important problem worthy of further exploration.

\paragraph{\rm\textbf{Acknowledgements}} This work is supported by GRF 14205719, TRS T41-603/20-R, Centre for Perceptual and Interactive Intelligence, and CUHK Interdisciplinary AI Research Institute. The authors would like to thank the valuable suggestions and comments by Xiaoyang Guo.

\clearpage
% !TEX root = ./supp.tex
\begin{center}
    \Large
    \textbf{Appendix}
\end{center}
\setcounter{section}{0}
\section{Supplementary Results}
\subsection{Detection Performance on the Test Set}
Due to the lack of pose information on the test set, we do not provide related results in the main paper. Here we show the results of our pose-free version in Tab.~\ref{tab: kitti_test}. We can observe conclusions similar to those on the validation set. Our method shows obvious superiority over previous methods even without precise pose information. Furthermore, we can expect a more significant improvement if ego-motion is available. We will also attempt to extend our method to other datasets that satisfy this requirement, such as nuScenes and Waymo\footnote{A simple extension version is presented in \cite{mvfcos3d++}.}.
\begin{table*}
\scriptsize
\caption{$AP_{40}$ results on the KITTI test set.}
\begin{center}
\setlength{\tabcolsep}{2.5mm}{
    \begin{tabular}{c|c|c|c|c|c|c|c}
    \hline
     \multirow{2}*{Methods} & \multirow{2}*{Venue} & \multicolumn{3}{c|}{AP$_{3D}$ IoU$\ge 0.7$ (\%)} & \multicolumn{3}{c}{AP$_{BEV}$ IoU$\ge 0.7$ (\%)}\\
    \cline{3-8}
    ~ & ~ & Easy & Mod. & Hard & Easy & Mod. & Hard\\
    \hline
    MonoDIS~\cite{MonoDIS} & ICCV 2019 & 10.37 & 7.94 & 6.40 & 17.23 & 13.19 & 11.12 \\
    M3D-RPN~\cite{M3D-RPN} & ICCV 2019 & 14.76 & 9.71 &	7.42 & 21.02 & 13.67 & 10.23 \\
    D4LCN~\cite{D4LCN} & CVPR 2020 & 16.65 & 11.72 & 9.51 & 22.51 &	16.02 &	12.55 \\
    MonoPair~\cite{MonoPair} & CVPR 2020 & 13.04 & 9.99 & 8.65 & 19.28 & 14.83 & 12.89 \\
    SMOKE~\cite{smoke} & CVPRW 2020 & 14.03 & 9.76 & 7.84 & 20.83 & 14.49 & 12.75 \\
    PatchNet~\cite{patchnet} & ECCV 2020 & 15.68 & 11.12 & 10.17 & 22.97 & 16.86 & 14.97 \\
    RTM3D~\cite{RTM3D} & ECCV 2020 & 14.41 & 10.34 & 8.77 & 19.17 & 14.20 & 11.99 \\
    IAFA~\cite{iafa} & ECCV 2020 & 17.81 & 12.01 & 10.61 & 25.88 & 17.88 & 15.35 \\
    MoVi3D~\cite{MoVi3D} & ECCV 2020 & 15.19 & 10.90 & 9.26  & 22.76 & 17.03 & 14.85 \\ 
    MonoDLE~\cite{monodle} & CVPR 2021 & 17.23 & 12.26 & 10.29 & 24.79 & 18.89 & 16.00 \\ 
    CaDDN~\cite{CaDDN} & CVPR 2021 & 19.17 & 13.41 & 11.46 & 27.94 & 18.91 & 17.19 \\
    MonoFlex~\cite{monoflex} & CVPR 2021 & 19.94 & 13.89 & 12.07 & 28.23 & 19.75 & 16.89 \\ 
    MonoRCNN~\cite{monorcnn} & ICCV 2021 & 18.36 & 12.65 & 10.03 & 25.48 & 18.11 & 14.10 \\ 
    GUPNet~\cite{gupnet} & ICCV 2021 & 20.11 & 14.20 & 11.77 & - & -  & - \\ 
    DFR-Net~\cite{DFR-Net} & ICCV 2021 & 19.40 & 13.63 & 10.35 & 28.17 & 19.17 & 14.84 \\
    \hline
    Kinematic3D~\cite{Kinematic3D} & ECCV 2020 & 19.07 & 12.72 & 9.17  & 26.69 & 17.52 & 13.10 \\
    \hline
    DfM w/o pose & ECCV 2022 & \textbf{22.94} & \textbf{16.82} & \textbf{14.65} & \textbf{31.71} & \textbf{22.89} & \textbf{19.97} \\
    \hline
    \end{tabular}
}
\end{center}
\label{tab: kitti_test}
\end{table*}

\subsection{Detection Performance of Other Classes}
Considering the limited samples of pedestrians and cyclists on KITTI, its performance is empirically unstable. So we mainly compare the detection performance of cars previously. Here, we also provide related results in Tab.~\ref{tab: kitti_ped_cyc} for reference. It can be seen that our method also achieves competitive results, especially on the detection of cyclists. For the detection of pedestrians, our method is only a little inferior to GUPNet~\cite{gupnet}. We suspect the reason is that the detection of small objects can be hard for BEV-based methods. From this perspective, our method achieves better performance than CaDDN, which follows a similar detection pipeline.
\begin{table*}
\scriptsize
\caption{$AP_{40}$ results of other classes on the KITTI test set.}
\begin{center}
\setlength{\tabcolsep}{2.5mm}{
    \begin{tabular}{c|c|c|c|c|c|c|c}
    \hline
     \multirow{2}*{Methods} & \multirow{2}*{Venue} & \multicolumn{3}{c|}{Ped@AP$_{3D}$ IoU$\ge 0.5$ (\%)} & \multicolumn{3}{c}{Cyc@AP$_{3D}$ IoU$\ge 0.5$ (\%)}\\
    \cline{3-8}
    ~ & ~ & Easy & Mod. & Hard & Easy & Mod. & Hard\\
    \hline
    M3D-RPN~\cite{M3D-RPN} & ICCV 2019 & 4.92 & 3.48 & 2.94 & 0.94 & 0.65 & 0.47 \\
    D4LCN~\cite{D4LCN} & CVPR 2020 & 4.55 & 3.42 & 2.83 & 2.45 & 1.67 & 1.36 \\
    MonoPair~\cite{MonoPair} & CVPR 2020 & 10.02 & 6.68 & 5.53 & 3.79 & 2.12 & 1.83 \\
    MoVi3D~\cite{MoVi3D} & ECCV 2020 & 8.99 & 5.44 & 4.57 & 1.08 & 0.63 & 0.70 \\ 
    MonoDLE~\cite{monodle} & CVPR 2021 & 9.64 & 6.55 & 5.44 & 4.59 & 2.66 & 2.45 \\ 
    CaDDN~\cite{CaDDN} & CVPR 2021 & 12.87 & 8.14 & 6.76 & 7.00 & 3.41 & 3.30 \\
    MonoFlex~\cite{monoflex} & CVPR 2021 & 9.43 & 6.31 & 5.26 & 4.17 & 2.35 & 2.04 \\ 
    GUPNet~\cite{gupnet} & ICCV 2021 & \textbf{14.72} & \textbf{9.53} & \textbf{7.87} & 4.18 & 2.65 & 2.09 \\ 
    DFR-Net~\cite{DFR-Net} & ICCV 2021 & 6.09 & 3.62 & 3.39 & 5.69 & 3.58 & 3.10 \\
    \hline
    Kinematic3D~\cite{Kinematic3D} & ECCV 2020 & - & - & -  & - & - & - \\
    \hline
    DfM w/o pose & ECCV 2022 & 13.70 & 8.71 & 7.32 & \textbf{8.98} & \textbf{5.75} & \textbf{4.88} \\
    \hline
    \end{tabular}
}
\end{center}
\label{tab: kitti_ped_cyc}
\end{table*}

\subsection{Latency of Constructing Cost Volume}
In the main paper, we mentioned that although cost volume construction becomes more complicated than that in the binocular system, it is overall achieved with matrix multiplication. The additional complexity increases the latency of this process from 0.003s to 0.012s, which can be ignored for the overall inference latency of 0.32s. Although our framework does not achieve real-time efficiency, it has performed better than similar baselines such as CaDDN (0.63s) and Pseudo-LiDAR based methods (about 0.4s). In addition, we can reduce the number of candidate depth levels to optimize the network efficiency while affecting little performance. We will also improve our framework in this aspect in the future.
\begin{table*}
\scriptsize
\caption{Ablation study for location-aware monocular compensation.}
\begin{center}
\setlength{\tabcolsep}{3mm}{
    \begin{tabular}{c|c|c|c|c|c|c}
    \hline
    \multirow{2}*{Methods} & \multicolumn{3}{c|}{AP$_{3D}$ IOU$\ge 0.7$} & \multicolumn{3}{c}{AP$_{BEV}$ IOU$\ge 0.7$}\\
    \cline{2-7}
    ~ & Easy & Mod. & Hard & Easy & Mod. & Hard\\
    \hline
    stereo baseline & 21.47 & 15.32 & 13.83 & 29.22 & 21.22 & 19.51\\
    \hline
    w/ shared weights & 22.92 & 15.99 & 13.85 & 31.31 & 23.00 & 20.22\\
    group-wise fusion & 23.49 & 16.52 & 14.38 & 33.00 & 23.91 & 21.06\\
    point-wise fusion & 26.61 & 18.82 & 16.47 & 36.16 & 26.09 & 23.17\\
    \hline
    \end{tabular}
}
\end{center}
\label{tab: ablation_mono_comp}
\end{table*}
\subsection{Supplementary Ablation Studies}
\noindent\textbf{Alternative Monocular Compensation Methods}\quad
We also attempt alternative methods to fuse monocular and stereo features. The final version in the main paper is both interpretable and effective. To have a more comprehensive comparison, we also show the results of other alternative designs for monocular compensation in Tab.~\ref{tab: ablation_mono_comp}. First, we use a simple convolution layer to directly compress these two feature volumes to one, \emph{i.e.}, compress from $2D$ channels to $D$. This implementation is simple while it essentially uses shared weights to aggregate these two volumes across the entire scene. As analyzed in the paper, we need to fuse adaptively because different locations can rely on monocular or stereo estimation differently. Then we attempt to use group-wise convolution to achieve this. Finally, our final version, generating a point-wise weight first and then using it to guide the fusion, is the most effective design. It is also in line with our theoretical analysis.

\begin{table*}
\scriptsize
\caption{Ablation study for depth loss design.}
\begin{center}
\setlength{\tabcolsep}{3mm}{
    \begin{tabular}{c|c|c|c|c|c|c}
    \hline
    \multirow{2}*{Methods} & \multicolumn{3}{c|}{AP$_{3D}$ IOU$\ge 0.7$} & \multicolumn{3}{c}{AP$_{BEV}$ IOU$\ge 0.7$}\\
    \cline{2-7}
    ~ & Easy & Mod. & Hard & Easy & Mod. & Hard\\
    \hline
    Mono Only w/ CE & 20.06 & 15.30 & 14.05 & 27.84 & 21.78 & 19.96 \\
    Stereo Only w/ CE & 21.47 & 15.32 & 13.83 & 29.22 & 21.22 & 19.51\\
    \hline
    Mono+Stereo w/ CE & 26.61 & 18.82 & 16.47 & 36.16 & 26.09 & 23.17\\
    \hline
    focal w/ gamma=2 & 27.27 & 18.76 & 16.55 & 35.29 & 25.08 & 22.11\\
    balanced w/ fg:bg=5:1 & 27.40 & 19.11 & 16.58 & 36.28 & 26.18 & 23.09\\
    balanced + focal & 29.27 & 20.22 & 17.46 & 38.60 & 27.13 & 24.05\\
    \hline
    \end{tabular}
}
\end{center}
\label{tab: ablation_depth_loss}
\end{table*}

\begin{table*}
\scriptsize
\caption{Depth estimation errors when using different loss designs. Err. Med. denotes the average median of depth errors and other metrics evaluate the ratio of points with errors larger than a specific threshold. Foreground (Fg in the table) metrics are evaluated by averaging object-level results. Objects with less than 5 ground-truth LiDAR points are ignored.}
\begin{center}
\setlength{\tabcolsep}{2mm}{
    \begin{tabular}{c|c|c|c|c|c}
    \hline
     \multirow{2}*{Methods} & \multicolumn{1}{c|}{Err. Med.$\downarrow$} & \multicolumn{1}{c|}{$>$0.2m$\downarrow$} & \multicolumn{1}{c|}{$>$0.4m$\downarrow$} & \multicolumn{1}{c|}{$>$0.8m$\downarrow$} & \multicolumn{1}{c}{$>$1.6m$\downarrow$}\\
    \cline{2-6}
    ~ & Fg/All (m) & Fg/All (\%) & Fg/All (\%) & Fg/All (\%) & Fg/All (\%)\\
    \hline
    Mono Only w/ CE & 5.86/1.15 & 91.2/75.6 & 83.9/64.4 & 74.6/53.1 & 65.8/43.3 \\
    Stereo Only w/ CE & 3.33/0.58 & 88.6/68.5 & 79.3/52.4 & 66.6/36.0 & 51.9/23.0 \\
    \hline
    Mono+Stereo w/ CE & 2.60/0.48 & 86.3/66.8 & 75.0/50.2 & 60.0/33.4 & 43.9/20.7 \\
    \hline
    focal w/ gamma=2 & 2.59/0.48 & 86.3/67.0 & 75.2/50.1 & 60.4/33.2 & 44.2/20.6 \\
    balanced w/ fg:bg=5:1 & 2.12/0.51 & 83.2/67.7 & 70.2/51.3 & 53.6/34.6 & 35.7/21.7 \\
    balanced + focal & 2.09/0.50 & 82.8/67.2 & 69.7/50.9 & 53.1/34.2 & 35.4/21.3 \\
    \hline
    \end{tabular}
}
\end{center}
\label{tab: ablation_depth_error}
\end{table*}

\noindent\textbf{Design of Depth Loss}\quad
Our baseline uses cross-entropy loss for depth supervision. Since our target is 3D object detection, we should pay more attention to foreground points. Therefore, following CaDDN~\cite{CaDDN}, we use focal design and balanced weights to facilitate the depth estimation from this aspect. We show their effectiveness in Tab.~\ref{tab: ablation_depth_loss} and \ref{tab: ablation_depth_error}. To have a more intuitive comparison, we also show related results of monocular and stereo only baselines. We can see these designs tailored to depth contribute a lot to the final performance improvement, which further shows the crucial role of depth estimation in monocular 3D detection.

\begin{table*}
\scriptsize
\caption{Our baseline performs much worse than its binocular counterpart. The key is the accuracy of depth distribution.}
\begin{center}
\setlength{\tabcolsep}{3mm}{
    \begin{tabular}{c|c|c|c|c|c|c}
    \hline
    \multirow{2}*{Methods} & \multicolumn{3}{c|}{AP$_{3D}$ IOU$\ge 0.7$} & \multicolumn{3}{c}{AP$_{BEV}$ IOU$\ge 0.7$}\\
    \cline{2-7}
    ~ & Easy & Mod. & Hard & Easy & Mod. & Hard\\
    \hline
    Binocular Baseline & 80.62 & 61.88 & 54.92 & 90.26 & 73.63 & 66.24\\
    w/ gt depth dist. & 85.41 & 70.07 & 62.96 & 93.82 & 82.24 & 74.89\\
    \hline
    DfM Baseline & 17.41 & 12.93 & 11.60 & 24.78 & 18.21 & 16.06\\
    w/ gt depth dist. & 76.70 & 63.01 & 55.74 & 87.47 & 76.62 & 69.09\\
    \hline
    \end{tabular}
}
\end{center}
\label{tab: motivating_study}
\end{table*}

\subsection{Oracle Analysis for Baseline Model}
When we build our baseline framework at the beginning (w/o data augmentation and monocular compensation), it turns out that the detection performance drops precipitously compared to the binocular baseline counterpart. However, if we replace the predicted depth distribution $\hat{D}_P$ with its target $D_P$, our baseline can be directly lifted to a level comparable with the binocular case. Although this assumption is a little idealistic, it still indicates that the key problem of this large gap is the accuracy of depth estimation. Therefore, we focus on improving the depth-from-motion component in the main paper and propose two effective designs.

\begin{figure}
    \centering
    \includegraphics[width=1.0\textwidth]{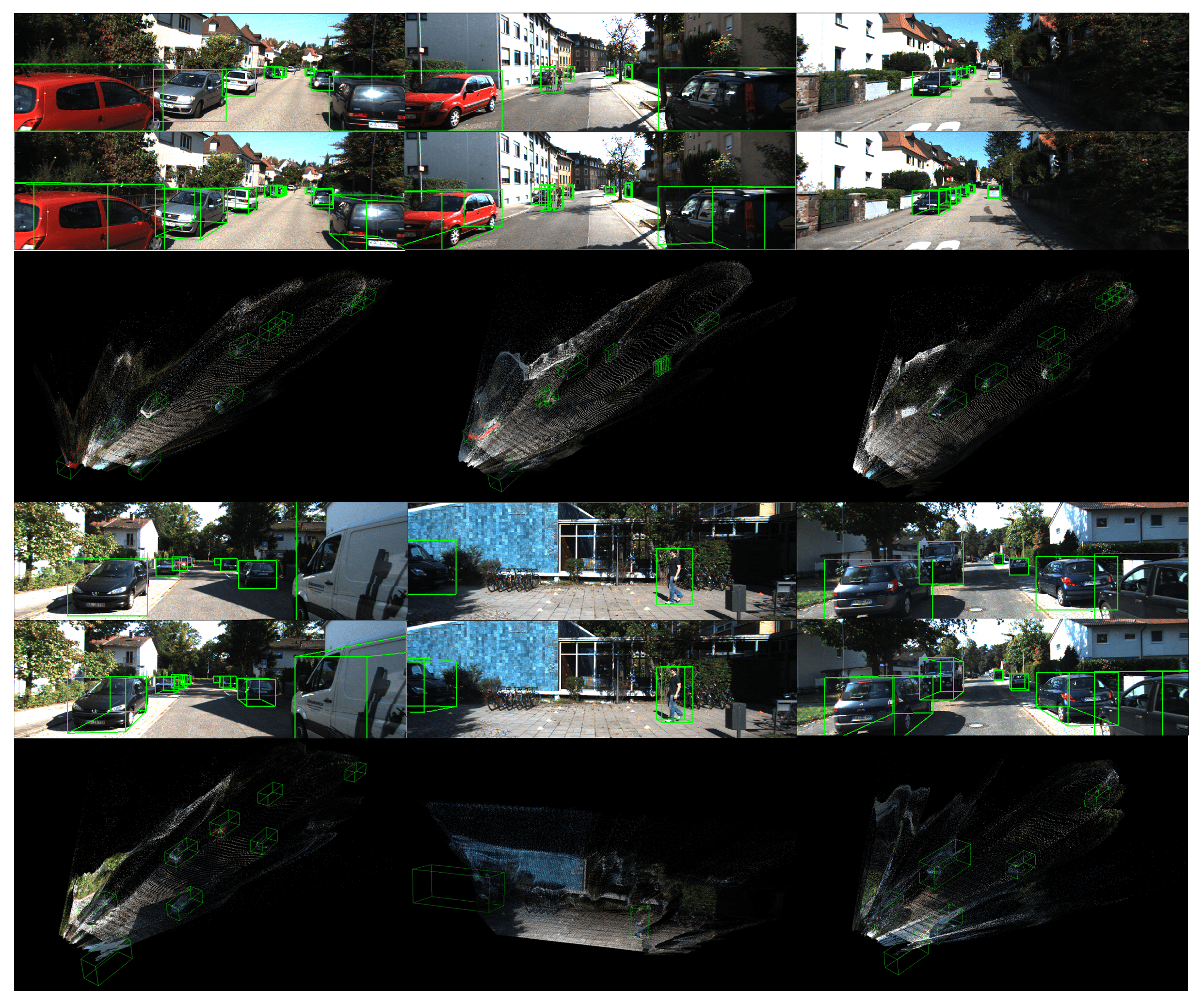}
    \caption{Qualitative detection results from the perspective view and 3D view.}
    \label{fig:qualitative_supp}
\end{figure}

\subsection{Qualitative Results}
We show detection results qualitatively in Fig.~\ref{fig:qualitative_supp}. For each sample, we visualize 2D detection and 3D detection results from the perspective view on the first two rows and plot 3D detection results in the 3D view on the third row. For the perspective view, we also reconstruct the point clouds with our estimated depth and paint them with corresponding colors. Please see qualitative results for 3D detection from consecutive frames in the supplementary demo video.

\begin{figure}
    \centering
    \includegraphics[width=1.0\textwidth]{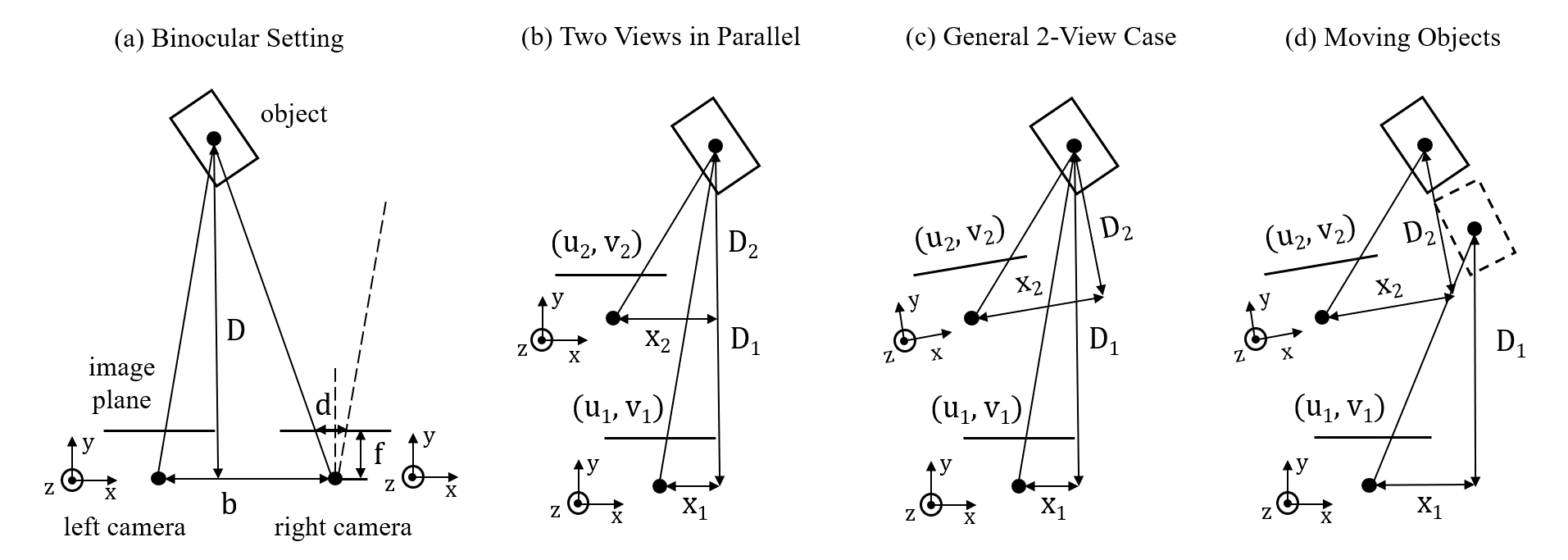}
    \caption{Multi-view geometry for object depth estimation in the (a) binocular, (b) parallel two-view, (c) general two-view system and (d) that for moving objects.}
    \label{fig:geometry}
\end{figure}
\section{Theoretical Analysis for General Two-View Cases}
We have discussed the geometry relationship in different two-view cases in the main paper, especially the two simplest cases. Although the cases with ego-motion and object motions are not important for the basic conclusion and our technical design, we still provide a basic analysis here for integration. It can also provide guidance for future work in this direction.
\subsection{General Two-View Case}
Following the basic analysis for the binocular system and two-parallel-view case, we extend the geometry analysis to the most general one without considering the motions of target objects: The pose transformation between two views consists of both translation and rotation (Fig.~\ref{fig:geometry}-(c)). Similar to the analysis for two parallel views, what we have are two projection relationships and the pose transformation:
\begin{equation}
    \left(
    \begin{matrix}
    u_1 \\
    v_1 \\
    1
    \end{matrix}
    \right)D_1 = \left(
    \begin{matrix}
    f & 0 & c_u \\
    0 & f & c_v \\
    0 & 0 & 1
    \end{matrix}
    \right)\left(
    \begin{matrix}
    x_1 \\
    y_1 \\
    D_1
    \end{matrix}
    \right),
\end{equation}

\begin{equation}
    \left(
    \begin{matrix}
    u_2 \\
    v_2 \\
    1
    \end{matrix}
    \right)D_2 = \left(
    \begin{matrix}
    f & 0 & c_u \\
    0 & f & c_v \\
    0 & 0 & 1
    \end{matrix}
    \right)\left(
    \begin{matrix}
    x_2 \\
    y_2 \\
    D_2
    \end{matrix}
    \right),
\end{equation}

\begin{equation}
    \left(
    \begin{matrix}
    x_2 \\
    y_2 \\
    D_2
    \end{matrix}
    \right) = 
    R_{3\times 3}\left(
    \begin{matrix}
    x_1 \\
    y_1 \\
    D_1\end{matrix}
    \right)+t_{3\times 1}.
\end{equation}
Represent $x_1, x_2, y_1, y_2$ with $D_1, D_2$, and substitute them in the transformation equation, and we can derive $D_1$ and $D_2$:
\begin{equation}
\begin{aligned}
   \Rightarrow D_2 &= (r_{31}\frac{u_1-c_u}{f}+r_{32}\frac{v_1-c_v}{f}+r_{33})D_1+t_{3}\\
   &\triangleq A_3 D_1+B_3,
\end{aligned}
\label{eqn:c_d2}
\end{equation}
where $r_{ij}$ denotes the i-th row, j-th column element of the rotation matrix $R$. Similarly, we can define:
\begin{equation}
    r_{11}\frac{u_1-c_u}{f}+r_{12}\frac{v_1-c_v}{f}+r_{13} \triangleq A_1, t_{1} \triangleq B_1
\end{equation}
\begin{equation}
    r_{21}\frac{u_1-c_u}{f}+r_{22}\frac{v_1-c_v}{f}+r_{23} \triangleq A_2, t_{2} \triangleq B_2
\end{equation}

Then:
\begin{equation}
   D_1 = (B_1-\frac{u_2-c_u}{f}B_3) / (\frac{u_2-c_u}{f}A_3-A_1)
\label{eqn:c_d1u}
\end{equation}
\begin{equation}
   D_1 = (B_2-\frac{v_2-c_v}{f}B_3) / (\frac{v_2-c_v}{f}A_3-A_2)
\label{eqn:c_d1v}
\end{equation}

When there is no rotation, setting $R$ to the identity matrix, it can also be reduced to the case with two views in parallel.

For this most complicated relationship, we can also understand it from the previous two cases. We first re-written Eqn.~\ref{eqn:c_d1u} as follows:
\begin{equation}
   D_1 = \frac{f(t_1-\frac{u_2-c_u}{f}t_3)}{\left(
    \begin{matrix}
    a_{31} a_{32} a_{33}
    \end{matrix}
    \right)\left(
    \begin{matrix}
    \frac{u_1-c_u}{f} \\
    \frac{v_1-c_v}{f} \\
    1\end{matrix}
    \right)(u_2-c_u)-\left(\begin{matrix}
    a_{11} a_{12} a_{13}
    \end{matrix}\right)\left(
    \begin{matrix}
    \frac{u_1-c_u}{f} \\
    \frac{v_1-c_v}{f} \\
    1\end{matrix}
    \right)f}
    \label{eqn:general-two-view}
\end{equation}

The numerator is the same with Eqn. 3 in the main paper while the denominator is coupled with some rotations. Here we can also substitute $\frac{u_1-c_u}{f}$ and $\frac{v_1-c_v}{f}$ with $\frac{x_1}{D}$ and $\frac{y_1}{D}$ (both correspond to rotations).

After primarily interpreting the result, let us recap Eqn.~\ref{eqn:c_d1u}, which is more clear for implementation. Here, if we would like to estimate depth directly, we need to predict $u_1, u_2, v_1$, and other values are constant given by the dataset. We can further transform the prediction of $u_2$ to $u_1+\Delta u$ to simplify the learning target and turn to address the correspondence problem. However, the prediction of $u_1$ and $v_1$ can also be inaccurate, so we can first use ground truths (target values of $u_1$ and $v_1$) to learn $\Delta u$ to observe whether it can converge or not. It turns out that even the task has been simplified a lot, it is still quite difficult from our preliminary experimental attempts. This is what we mentioned in the main paper: In this case, disparity computation involves several rotation coefficients and additional dimensions of absolute position $v_1$. The cumulative errors caused by the entanglement of multiple estimations make the direct derivation intractable.

\subsection{Moving Objects}
Up to now, all the formulation assumes the object is static. However, there are many moving objects in the open world. Next, we will discuss what will happen if we consider the moving objects.

Let us consider the 3D center of a car (Fig.~\ref{fig:geometry}-(d)): it can only drift (both in 3D and 2D) when the car has a translation. Rotation does not affect its 3D location and thus does not affect its 2D projection. Therefore, for object centers, the only difference in the previous relationship (Eq.~\ref{eqn:general-two-view}) is just that the object translation should be added into the translation vector $t_{3\times 1}$. However, this can be different for other 3D points. For example, the points on the object surface can rotate with the object's rotation, which can be hard to formulate with our current modeling.

The basic analysis shows that moving objects can involve local warping to monocular images, in contrast to global warping caused by view change. Due to the complexity of different objects' motion and the domain gap between the 3D targets and 2D inputs, it is hard to directly estimate motion from only a pair of images, not to mention involving the estimation errors in the direct computation of depth.

From the perspective of our framework in the main paper, a promising direction is to model the local warping when constructing stereo cost volume and attempt to remove this factor for stereo matching. More annotations such as complete tracklets may be required for better performance.

\section{Implementation Details}
In the main paper, we have introduced our overall framework and detailed our proposed two key components. This supplemental section elaborates on the specific network architectures of other basic modules and presents the design related to auxiliary tasks except for 3D detection.

Our framework is motivated by DSGN~\cite{dsgn} and LIGA-Stereo~\cite{liga}. We will also release our code afterward for reproducing our experiments and showing these details more conveniently.

\subsection{Network Architecture}
\noindent\textbf{2D Feature Extraction}\quad Given the input image-pair $(I_t, I_{t-\delta t})$, we use a shared 2D backbone to extract their features $(\mathcal{F}_t, \mathcal{F}_{t-\delta t})$. The backbone is a modified ResNet34~\cite{ResNet} with the channels of $conv2-5$ being set to \{64,128,128,128\}. The design of SPP~\cite{spp} module follows DSGN~\cite{dsgn}. We append a small U-Net~\cite{unet} to upsample these SPP features to get the full resolution $\mathcal{F}_t$ for high-quality stereo matching while use 2-layer convolution to extract the semantic feature $\mathcal{F}_{sem}$~\cite{liga}. The final number of channels are set to 32 for both $\mathcal{F}_t$ and $\mathcal{F}_{sem}$.

\noindent\textbf{2D Detection Head}\quad We construct five-level FPN~\cite{FPN} by appending multiple stride-2 convolution layers on the SPP feature of frame $t$. Then we attach a 2D detection head for each level following ATSS~\cite{atss}. Each position only has one anchor box and the anchor box sizes on
each level are set to \{32, 64, 128, 256, 512\}.

\noindent\textbf{2.5D Backbone}\quad After constructing the monocular and stereo cost volume, we filter each with a 3D residual block and a 3D hourglass network separately. The residual block consists of two 3D convolution layers and a skip connection as the basic block in ResNet. The 3D hourglass network downsamples the 3D feature with two stride-2 layers and then upsamples them with skip connections. We set the 3D kernel size to 3$\times$3$\times$3 by default. While all these operations are 3D convolutions, we call this component as 2.5D backbone because the spatial quantization is based on the 2.5D coordinates, \emph{i.e.}, following the plane-sweep approach, which is different from the voxelization in the 3D space.

\noindent\textbf{3D Backbone and 3D Head}\quad With the fused stereo feature, the depth head applies a simple 3D convolution layer followed by softmax to predict the depth distribution. We further apply the outer product to the semantic feature $F_{sem}$ and the depth probability volume $D_P$, and combine it with the stereo feature for sampling the voxel features used for subsequent 3D detection. The voxel feature is then filtered with a 3D convolution layer and downsampled along the height axis. We transform this feature by merging its height and feature dimension to get the bird-eye-view (BEV) feature. A 2D hourglass network with 2-layer downsampling and upsampling is applied on top to get the input of 3D heads. Finally, we append two layers for the classification and regression branch separately and use one layer for each task: classification, direction classification and regression. We follow the rotation encoding scheme in SECOND~\cite{SECOND}, and use kernel size 3$\times$3 by default for all the layers except the final one for direction classification.

\subsection{Training Loss}
As mentioned in the paper, the training loss is composed of depth loss $\mathcal{L}_{depth}$, 2D detection loss $\mathcal{L}_{2D}$ and 3D detection loss $\mathcal{L}_{3D}$. Our training loss follows LIGA-Stereo~\cite{liga} while having minor modifications on depth loss. To make the paper self-contained, we briefly introduce them as follows.

First, the baseline cross-entropy depth loss is:
\begin{equation}
    \mathcal{L}_{depth} = \frac{1}{N_{gt}}\sum_{u,v}\sum_{w}\Big[-max(1-\frac{|d^*-d(w)|}{\Delta d}, 0) log\mathcal{D}_{P}(u,v,w)\Big],
\end{equation}
where $N_{gt}$ is the number of valid pixels with depth ground truth $d^*$, $u,v,w$ denotes the position in the stereo volume, $\Delta d$ is the divided depth interval as in the main paper.

We upgrade it with balanced weights and focal design to make it more concentrated on foreground points. The foreground and background weight of depth estimation is set to 5 and 1, and $\gamma$ is set to 2 in the focal loss.

2D detection loss $\mathcal{L}_{2D}$ consists of three parts: focal loss for classification $\mathcal{L}^{cls}_{2D}$, GIoU loss for localization $\mathcal{L}^{GIoU}_{2D}$ and cross-entropy loss for centerness $\mathcal{L}^{ct}_{2D}$. The weights of them are set to 1.0, 2.0, 1.0 respectively.

3D detection loss $\mathcal{L}_{3D}$ has four components: focal loss for 3D classification $\mathcal{L}^{cls}_{3D}$, regression L1 loss $\mathcal{L}^{reg}_{3D}$ and IoU loss $\mathcal{L}^{IoU}_{3D}$ for localization, and cross-entropy loss for direction classification $\mathcal{L}^{dir}_{3D}$. Except for IoU loss, the others are devised following SECOND~\cite{SECOND}. The IoU loss is defined as the average rotated IoU loss between the predicted boxes and ground truth boxes. The weights of them are set to 1.0, 0.5, 1.0, 0.2. In addition, we keep the original imitation loss $\mathcal{L}_{im}$ in LIGA-Stereo~\cite{liga} to learn better geometric information. We keep its weight to 1.0 and obtain a little performance gain of about 1 AP in our baseline.

\subsection{View Synthesis in Pose-Free DfM}
When computing the self-supervised loss for pose learning in the pose-free DfM, the main paper mentioned that we need to synthesize the frame $t$ with frame $t-\delta t$. Here we detail the synthesis procedure.

With the dense depth estimation $\hat{D}_t$, we can obtain a stereo grid by reprojecting the 2D grid of frame $t$ with the intrinsic matrix. Then we warp these positions to the frame $t-\delta t$ with the predicted pose $(\mathbf{t},\mathbf{q})$ and project them to the image plane to sample corresponding pixels. The sampled result is the expected synthesized $I_{t-\delta t\rightarrow t}$. Note that we also apply the pose-based warping in the canonical space similar to the construction of cost volume in this procedure.

\section{Supplementary Video}
We attach a video in the supplementary material. This video first combs out our method's general logic and specific content so that reviewers can understand or recap it quickly. The end shows some demo videos of the 3D detection results predicted by our model, from the perspective view and 3D view, respectively. It supplements the main paper on the qualitative results of consecutive-frame images. The video is compressed in the supplementary file. Please see the full version provided at \url{https://github.com/Tai-Wang/Depth-from-Motion}.
\bibliographystyle{splncs04}
\bibliography{egbib}
\end{document}